\documentclass{article}

\usepackage{PRIMEarxiv}

\usepackage[utf8]{inputenc} 

\usepackage[T1]{fontenc}    
\usepackage{hyperref}  
\usepackage{url}  
\usepackage{float}
\usepackage{textcase}
\usepackage[labelfont=bf, textfont={bf,it}]{caption}
\usepackage{subcaption}
\usepackage[noend]{algpseudocode}
\usepackage{algorithm}
\usepackage{lineno}
\usepackage[font=large]{caption} 
\usepackage{newtxtext}

\usepackage{tikz}         
\usepackage{subcaption}   

\usepackage{caption}
\usepackage{indentfirst}

\usepackage{booktabs}       
\usepackage{amsfonts}       
\usepackage{nicefrac}       
\usepackage{microtype}      
\usepackage{lipsum}
\usepackage{multirow}
\usepackage[justification=centering]{caption}
\usepackage{indentfirst}
\usepackage{fancyhdr}
\usepackage{amsmath}
\usepackage{breqn}   
\graphicspath{{media/}} 

\title{
Toward a Universal Characterization of Sudoku Puzzle Difficulty 
}

\author{
  Arman Eisenkolb-Vaithyanathan \\
  Lynbrook High School \\
  San Jose\\
  \texttt{arman.eisenkolbvaithyanathan@gmail.com} \\
}
\makeatletter
\renewcommand{\maketitle}{
  \begin{center}
    {\LARGE\rmfamily \@title \par}
    \vskip 1em
    {\large\rmfamily
      \lineskip .5em
      \begin{tabular}[t]{c}
        \@author
      \end{tabular}\par}
    \vskip 1em
  \end{center}
}
\makeatother

\begin{document}
\raggedbottom
\vspace*{2cm} 

\maketitle

\begin{abstract}
\setlength\parindent{20pt}
In this paper we try to answer the question "What constitutes Sudoku difficulty rating across different Sudoku websites?"  Using two distinct methods that can both solve every Sudoku puzzle, I propose two new metrics to characterize Sudoku difficulty. The first method is based on converting a Sudoku puzzle into its corresponding Satisfiability (SAT) problem. The first proposed metric is derived from SAT Clause Length Distribution which captures the structural complexity of a Sudoku puzzle including the number given digits and cells they are in. The second method simulates human Sudoku solvers by intertwining four popular Sudoku strategies within a backtracking algorithm called Nishio. The second metric is computed by counting the number of times Sudoku strategies are applied within the backtracking iterations of a randomized Nishio. Using these two metrics, I analyze more than a thousand Sudoku puzzles across five popular websites to characterize every difficulty level in each website. I evaluate the relationship between the proposed metrics and website-labeled difficulty levels using Spearman’s rank correlation coefficient, finding strong correlations for 4 out of 5 websites. I construct a universal rating system using a simple, unsupervised classifier based on the two proposed metrics. This rating system is capable of classifying both individual puzzles and entire difficulty levels from the different Sudoku websites into three categories \textendash\  Universal Easy, Universal Medium, and Universal Hard \textendash\ thereby enabling consistent difficulty mapping across Sudoku websites. The experimental results show that for 4 out of 5 Sudoku websites, the universal classification aligns well with website-labeled difficulty levels. Finally, I present an algorithm that can be used by early Sudoku practitioners to solve Sudoku puzzles.
\end{abstract}

\section{Introduction}
\setlength\parindent{20pt}Sudoku is a widely popular logic-based puzzle enjoyed around the world. A Sudoku puzzle consists of a 9x9 grid of 81 cells, further divided into 9 smaller 3x3 grids, or commonly referred to as boxes (see Figure \ref{NyTimes Sudoku}). The objective is to fill the puzzle with the digits 1 through 9, such that each row, column, and box contains each digit without repetition. As a general term, rows, columns, and boxes will henceforth be called "sub-groups". Each puzzle begins with a set of pre-filled cells to provide a starting point. The puzzle is considered solved when all 81 cells are filled in accordance with these rules.

\setlength\parindent{20pt} There are tens if not hundreds of online Sudoku websites. Although all Sudoku puzzles are dictated by the same set of rules, not all puzzles are equally difficult. Indeed, the primary differentiator between websites are the difficulty levels and the puzzles within each difficulty level. Each site has their own way of defining Sudoku difficulty, and each site often has a different number of difficulty levels. For example, New York Times has three levels of difficulty - Easy, Medium, and Hard - while Sudoku.org.uk has four levels of difficulty - Gentle, Moderate, Tough, and Diabolical. The classification of Sudoku difficulty on each site is entirely done by the individual site. This paper analyzes Sudoku puzzles collected from 5 websites: New York Times \cite{NYT}, Sudoku.org.uk \cite{Sudoku.org.uk}, Extreme Sudoku \cite{Extreme_Sudoku}, Sudoku of the Day \cite{SOFTD}, and Sudoku of the Day UK \cite{SOFTDUK}.

\setlength\parindent{20pt} To understand the difficulty levels within each website and also compare them across websites, the first step is to characterize the difficulty of a Sudoku puzzle. To this end, I employ two distinctly different approaches. The first, purely computational, involves converting a Sudoku puzzle into the well-known boolean satisfiability problem (SAT). This allows the use of the characteristics of a SAT instance, specifically the length of clauses, to characterize the difficulty of Sudoku. 

\setlength\parindent{20pt}The second approach involves simulating the way a human solves a Sudoku puzzle. There are tens of Sudoku-solving strategies that people use to solve Sudoku puzzles. Ranging from easy to very complex, such strategies may involve sophisticated pattern identification on a Sudoku puzzle. To characterize Sudoku difficulty across a variety of human skill levels, I choose four Sudoku strategies that range from beginner to moderately sophisticated. While the application of these four strategies, alone, can solve some of the Sudoku puzzles on most websites, they do not solve all the puzzles on all websites. Therefore, for the second approach, I use a simple trial-and-error methodology in conjunction with the four human strategies. The trial and error implementation for Sudoku, known as \textit{Nishio}, with the four human strategies, can solve every Sudoku puzzle. This work simulates random humans solving Sudoku puzzles using the four human strategies within a randomized version of Nishio. Difficulty is characterized by counting the number of times Sudoku strategies are applied within Nishio by the simulated human solver. The average count within each difficulty level provides a metric that characterizes difficulty both within a website and across websites. Finally, I use the two proposed metrics to create a universal rating system built on a simple, unsupervised classifier to enable comparison across the five Sudoku websites and classification of unlabeled datasets.

\setlength\parindent{20pt}This paper is organized as follows. The next section outlines the key contributions of the work. Following that, a running example is introduced along with three core constructs common to all Sudoku puzzles.  This is followed by a section describing the encoding of Sudoku puzzle into SAT. Then  the Nishio interleaving Human Strategies method is presented, followed by the data sets used in this study. Next, results for both methods are reported, followed by a detailed analysis conducted both within individual sites and across different sites. I then present a universal classification of Sudoku difficulty.  Finally, the paper concludes with a proposed human solving algorithm derived from the study's findings.

\section{Key Contributions}
The main contributions of this paper are given below..
\begin{itemize}
    \item \textbf{Two Novel Difficulty metrics} I propose and evaluate two new metrics for characterizing Sudoku puzzle difficulty. The first, \textit{Clause Length Distribution}, is derived from SAT encodings of Sudoku puzzles. \textit{Nishio Human Cycles} are based on a simulation of human-like Sudoku solving using a randomized trail-and-error (Nishio) embedded with four Sudoku solving strategies. 
    \item \textbf{Percentage of Puzzles Solved by Human Strategies Alone} is an additional dimension through which the five Sudoku websites are analyzed.
    \item \textbf{A universal difficulty classification that can classify individual puzzles and different difficulty levels within each website}. Using the two proposed above, I build a universal classification of Sudoku difficulty with three categories. Each difficulty level from five major Sudoku websites is mapped onto one of these categories thereby enabling cross-website comparison and a standardization of seemingly disparate difficulty levels. 
    \item \textbf{A 1320 puzzle dataset} from five Sudoku websites with difficulty classification for each puzzle.

    \item \textbf{A method for early Sudoku practitioners} to solve everyday puzzles. 
    
\end{itemize}

Section \ref{related work}, titled Related Work, presents a detailed comparison between this work and previous studies.

\section{Running Example: Figure 1} \label{Running Example and Candidates}
\setlength\parindent{20pt}
Figure 1 shows two Sudoku puzzles released by the New York Times \cite{NYT} on July 23, 2024. These two puzzles will serve as the running example throughout this paper. Below, I describe the three core constructs of every Sudoku puzzle: given digits, numbered rows and columns, and candidates. 

\begin{figure}[t]
    \rule[1ex]{\textwidth}{0.1pt}
    \begin{subfigure}[b]{0.50\textwidth}
        \centering
        \includegraphics[width =\textwidth]{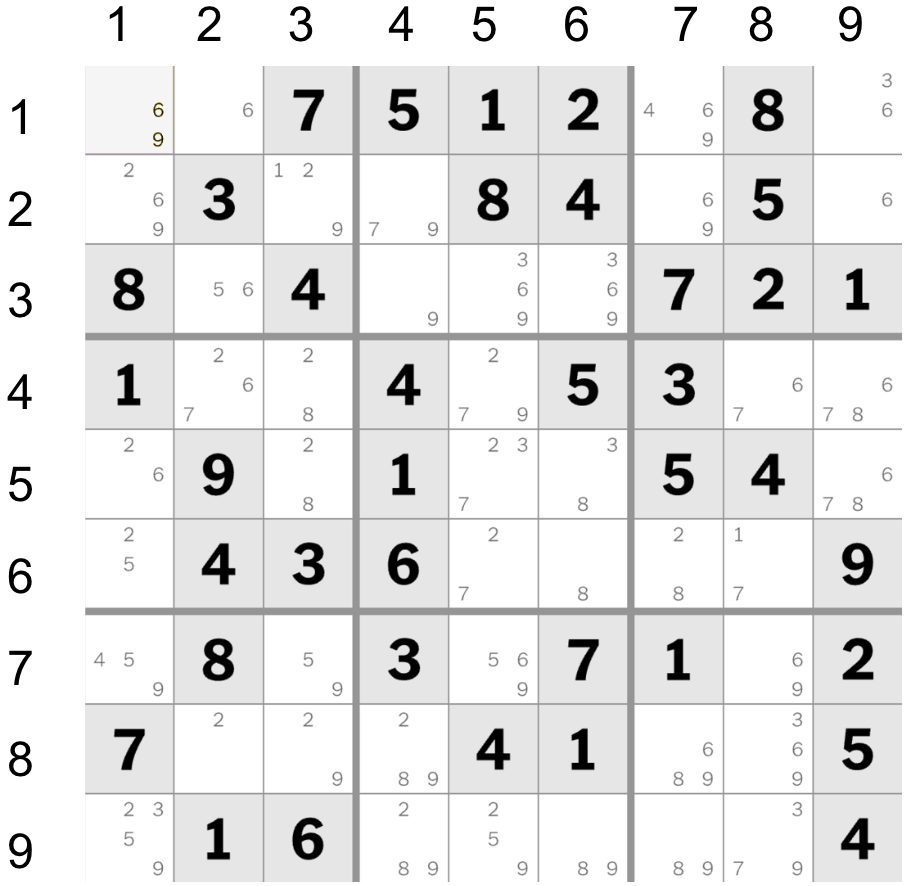}
        \caption{Easy}
        \label{Easy Sudoku}
    \end{subfigure}
    \hfill
    \begin{subfigure}[b]{0.47\textwidth}
        \centering
        \includegraphics[width =\textwidth]{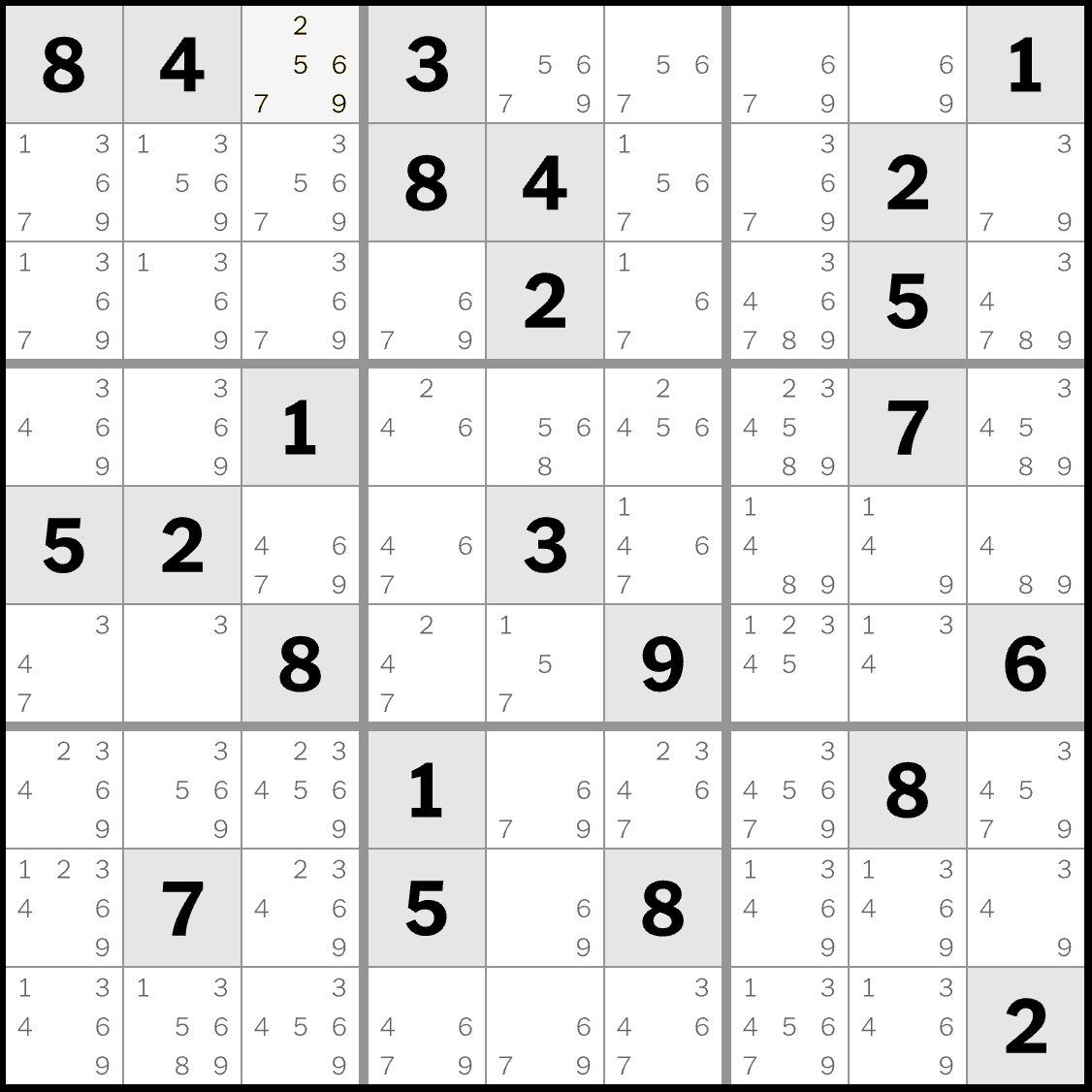}
        \caption{Medium}
        \label{Medium Sudoku}
    \end{subfigure}
    \caption{Two New York Times Sudoku Puzzles from July 23rd 2024 with candidates filled in \label{NyTimes Sudoku}}
    \rule[1ex]{\textwidth}{0.1pt}
\end{figure}

\paragraph{Given Digits} Each puzzle has pre-filled cells, denoted by large bold digits in gray squares in Figure \ref{NyTimes Sudoku}, and are known as "given digits". The human solver uses the given digits to then progress towards solving the puzzle.  

\paragraph{Numbered Rows and Columns} In Figure \ref{Easy Sudoku}, there is a digit on top of each column and to the left of each row. These digits, ranging from 1 to 9, represent the row and column numbers. A cell in any given Sudoku is represented by (x, y), where x is the row number and y is the column number. 

\paragraph{Candidates}
Candidates are the possible digits that can fill the empty cell which they are in.  In Figure \ref{NyTimes Sudoku}, candidates are denoted in empty cells by a set of digits in smaller gray font. To compute candidates \cite{solving_sudoku_by_hand}, apply Sudoku's rules to identify which digits (1–9) are not yet placed in the intersecting row, column, and box of each cell. As an example, cell (1, 7) in Figure \ref{Easy Sudoku}, consider row 1 - the row that intersects cell (1, 7); digits 7, 5, 1, 2, and 8 already appear, ruling those out as candidates for cell (1, 7). Further, column 7 contains digit 3, ruling it out as a candidate. The remaining digits, which do not appear in the row, column, and box intersecting cell (1, 7) are 4, 6, and 9 hereby making them candidates for cell (1, 7). The candidates for other cells are filled in similar fashion. 

\section{Encoding a Sudoku Puzzle as Satisfiability Problem}
\setlength\parindent{20pt} A boolean satisfiability problem, or SAT, is a fundamental concept in computer science which involves determining whether a given formula, with only boolean variables, has a solution. Each SAT is made up of clauses and literals. A clause is a disjunction of literals and each literal is a boolean variable. Once each literal is given a boolean value, a SAT problem is considered solved if all clauses are true. Encoding a Sudoku puzzle as a SAT problem allows using SAT solvers to find solutions to Sudoku puzzles efficiently. To accomplish this, the rules and constraints of Sudoku are translated into clauses and literals, and then SAT solvers are used to solve a Sudoku puzzle's SAT formulation. Further, it is possible to analyze properties of a SAT instance and its respective clauses and literals \cite{sat_hardness} to help characterize Sudoku difficulty.

\subsection{Converting a Sudoku Puzzle into SAT}
\setlength\parindent{20pt}The goal is to create a formula in conjunctive normal form (CNF), where the clauses are connected by the "AND" operator, and the literals within each clause are separated by the "OR" operator. Each literal represents a boolean variable or its negation with similar notation following the format used by Lynce and Ouaknine \cite{Sudoku-as-SAT}.  A literal \begin{math} C_{xyz} \end{math} is said to be true if cell (x, y) contains digit z.

There are four main constraints in Sudoku: cell, row, column, and box constraints. Row, column, and box are also referred to as "sub-groups". Each constraint has two types of clauses:

\begin{itemize}
    \item \textbf{At-least-one clauses}: 
    \begin{itemize}
        \item  Each cell contains at least one digit.
        \item Each row, column, and box contains every digit at least once.

    \end{itemize}
    \item \textbf{At-most-one clauses}: 
    \begin{itemize}
        \item  Each cell contains at most one digit 
        \item  Each row, column, and box contains every digit at most once.
    \end{itemize}
\end{itemize}

\subsubsection{At-Least-One Clauses} 

\paragraph{At-Least-One Cell Constraint Clauses} To ensure that at least one digit is present in each cell, I define an at-least-one clause for each cell. These clauses are expressed in equation \ref{eq:at-least-cell}, where (x, y) represents the coordinates of the cell and \begin{math}z \in\{1, 2, 3, 4, 5, 6, 7, 8, 9\}\end{math}.

\begin{equation} \label{eq:at-least-cell}
C_{xy1} \lor C_{xy2} \lor C_{xy3} \lor C_{xy4} \lor C_{xy5} \lor C_{xy6} \lor C_{xy7} \lor C_{xy8} \lor C_{xy9}
\end{equation}

\paragraph{At-Least-One Row Constraint Clauses} Each digit must appear at least once in each row. For every digit \begin{math}z \in\{1, 2, 3, 4, 5, 6, 7, 8, 9\}\end{math}, this condition is represented in equation \ref{eq:at-least-row}, where \begin{math}x \in\{1, 2, 3, 4, 5, 6, 7, 8, 9\}\end{math} represents every row number:
    \begin{equation} \label{eq:at-least-row}
    C_{x1z} \lor C_{x2z} \lor C_{x3z} \lor C_{x4z} \lor C_{x5z} \lor C_{x6z} \lor C_{x7z} \lor C_{x8z} \lor C_{x9z}
    \end{equation}
\paragraph{At-Least-One Column Constraint Clauses} Each digit must appear at least once in each column. For every digit \begin{math}z \in\{1...9\}\end{math}, this condition is represented in equation \ref{eq:at-least-col}, where \begin{math}y \in\{1...9\}\end{math} represents every column number:
    \begin{equation} \label{eq:at-least-col}
    C_{1yz} \lor C_{2yz} \lor C_{3yz} \lor C_{4yz} \lor C_{5yz} \lor C_{6yz} \lor C_{7yz} \lor C_{8yz} \lor C_{9yz}
    \end{equation}
\paragraph{At-Least-One Box Constraint Clauses} Each digit must appear at least once in each 3x3 box. For every digit \begin{math}z \in\{1...9\}\end{math} and \begin{math} x, y \in\{1,4,7\}\end{math}, this condition is represented in equation \ref{eq:at-least-box}:

\begin{equation} \label{eq:at-least-box}
    C_{xyz} \lor C_{x{(y+1)}z} \lor C_{x{(y+2)}z} \lor C_{(x+1)yz} \lor C_{(x+1){(y+1)}z} \lor C_{(x+1){(y+2)}z} \lor C_{(x+2)yz} \lor C_{(x+2){(y+1)}z} \lor C_{(x+2){(y+2)}z}
\end{equation}

 Equation \ref{eq:at-least-box}, when x = 1, y = 1, and z = 1, states that the digit 1 must appear at least once in the top left box. 

\setlength\parindent{20pt} There are 81 at-least-one clauses for each constraint combining to form a total of 324 clauses, with each clause containing 9 literals. A clause's length is determined by the number of literals it contains, and a clause with length $k$ is denoted as 1-in-$k$ SAT clause. So equations \ref{eq:at-least-cell}, \ref{eq:at-least-row}, \ref{eq:at-least-col}, and \ref{eq:at-least-box}, containing 9 literals each, are 1-in-9 SAT clauses.

\subsubsection{At-Most-One Clauses}\setlength\parindent{20pt}While the at-least-one clauses ensure that each digit appears at least once within each constraint (cell, row, column, box), they do not prevent the same digit from appearing multiple times within the same sub-group (or more than one digit appearing in the same cell). To enforce this rule, at-most-one clauses are used. These clauses enforce that no two digits can appear simultaneously in the same cell and the same digit can not appear twice in the same row, column, or box.

\paragraph{At-Most-One Cell Constraint Clauses}For each cell (x,y), the clauses in equations \ref{eq:at-most-cell}, \ref{eq:at-most-row}, \ref{eq:at-most-col}, and \ref{eq:at-most-box} ensure that no two digits can occupy the same cell. For example, the first clause states that if digit 1 appears in the cell, then digit 2 must not appear in the same cell. Expressing this relationship for all pairs of digits (e.g.,  1 and 2, 1 and 3,  1 and  4, and so on) results in:

\begin{equation} \label{eq:at-most-cell}
\begin{aligned}
\neg{C_{xy1}} \lor \neg{C_{xy2}}\\
\neg{C_{xy1}} \lor \neg{C_{xy3}}\\
...\hspace{2cm}\\
\neg{C_{xy7}} \lor \neg{C_{xy9}}\\
\neg{C_{xy8}} \lor \neg{C_{xy9}}\\
\end{aligned}  
\end{equation}
\setlength\parindent{20pt} In total there are \begin{math}{9 \choose 2} = 36\end{math} pairs of digits for each cell, and each pair gets its own clause ensuring that these two digits cannot appear together in the same cell. 36 clauses for each of the 81 cells results in a total of 2916 clauses. 

\setlength\parindent{20pt} Next, the at-most-one clauses guarantee that each digit can only appear once in each sub-group. 

\paragraph{At-Most-One Row Constraint Clauses} Each digit can appear at most once in each row. To enforce this, the clause in equation \ref{eq:at-most-row} ensures that no two cells in the same row can contain the same digit. For example, the first clause ensures that if digit z appears in cell (x,1), then it cannot appear in cell (x,2). Expressing this relationship for all pairs of cells in the row results in equation \ref{eq:at-most-row}:
\begin{equation} \label{eq:at-most-row}
\begin{aligned}
\neg{C_{x1z}} \lor \neg{C_{x2z}} \\
...\hspace{2cm}\\
\neg{C_{x8z}} \lor \neg{C_{x9z}} \\
\end{aligned} 
\end{equation}
\paragraph{At-Most-One Column Constraint Clauses} Similarly, each digit can appear at most once in each column. These clauses ensure that no two cells in the same column can contain the same digit. For example, the first clause ensures that if digit z appears in cell (1,y), then it cannot appear in cell (2,y). Expressing this relationship for all pairs of cells in the column results in equation \ref{eq:at-most-col}:
\begin{equation} \label{eq:at-most-col}
\begin{aligned}
\neg{C_{1yz}} \lor \neg{C_{2yz}}\\
...\hspace{2cm}\\
\neg{C_{8yz}} \lor \neg{C_{9yz}}\\
\end{aligned}
\end{equation}
\paragraph{At-Most-One Box Constraint Clauses} Each digit can appear at most once in each box. These clauses ensure that no two cells in the same 3x3 box contain the same digit. For example, the first clause ensures that if digit z appears in cell (x,y), then it cannot appear in cell (x,y+1). Expressing this relationship for all pairs of cells in a box, where \begin{math} x, y \in\{1,4,7\}\end{math}, results in equation \ref{eq:at-most-box}:
\begin{equation}\label{eq:at-most-box}
\begin{aligned}
\neg{C_{xyz}} \lor \neg{C_{x(y+1)z}}\hspace{1.75cm}\\
...\hspace{4.33cm}\\
\neg{C_{(x+2)(y+1)z}} \lor \neg{C_{(x+2)(y+2)z}}\\
\end{aligned}   
\end{equation}

\setlength\parindent{20pt} At-most-one clauses creates pairwise relationships between candidates in a cell and cells in a sub-group. Therefore, all at-most-one clauses are 1-in-2 SAT clauses. In total, the at-most-one clauses generate \begin{math}36 * 81 = 2916\end{math} 1-in-2 SAT clauses for each constraint.

\paragraph{Given Digits Clauses} The final set of clauses make a SAT formulation unique for any Sudoku puzzle and represent the given digits in each puzzle. These clauses are 1-in-1 SAT clauses, containing one literal each. Three such clauses for Figure \ref{Easy Sudoku} are outlined in equation \ref{eq: given digits}.

\begin{equation}\label{eq: given digits}
\begin{aligned}
{C_{137}}\\
{C_{145}}\\
{C_{151}}\\
\end{aligned}   
\end{equation}

\setlength\parindent{20pt} Equation \ref{eq: given digits} ensures that digit 7 goes in cell (1, 3), digit 5 goes in cell (1, 4) and digit 1 goes in cell (1, 5). These clauses are expressed for all given digits in the Sudoku resulting in 38 such clauses for Figure \ref{Easy Sudoku}. Similarly, there are 23 given digits clauses for Figure \ref{Medium Sudoku}.

\setlength\parindent{20pt} In total there are 2916 clauses for each of the four constraints. Adding the 324 clauses from the at-least-one clauses, the total number of clauses for a single empty Sudoku puzzle reaches 11988. The total number of clauses will vary based on the number of given digits. The number of clauses for Figure \ref{Easy Sudoku} is 12026. The method of encoding a Sudoku puzzle into a SAT problem outlined in this section is known as \textit{maximum encoding} \cite{Sudoku-as-SAT}. In the next section, we  explore how to use candidates to reduce the number of clauses for a single Sudoku puzzle by more than \begin{math} 90\% \end{math} in some cases.

\subsection{Using Candidates to Reduce the Number of Clauses }\label{candidate encoding}
\setlength\parindent{20pt}While the formulation in the previous section has sufficient clauses to solve a Sudoku as SAT, the number of clauses and literals can be reduced. To do this we can use the candidates, which are described in Section \ref{Running Example and Candidates}. This method of clause creation is known as \textit{minimum encoding} \cite{Sudoku-as-SAT}. Using this method, the number of literals can be reduced significantly. Let's take for instance cell (1,3) in figure \ref{Medium Sudoku}. The following, \begin{math} C_{131}, C_{133}, C_{134}, C_{138} \end{math}, will not show up as literals within the clauses since they are not candidates for cell (1,3).

\setlength\parindent{20pt}In Figure \ref{Medium Sudoku}, with maximum encoding, the clause for cell (1, 1) would consist of one literal for each digit, 1 - 9. Instead, using the candidates and the given digits, the cells (1, 1) and (1, 2) are skipped as they are pre-filled and the clause for cell (1, 3) will be shortened to a 1-in-5 SAT clause:
\begin{dmath}
C_{132} \lor C_{135} \lor C_{136} \lor C_{137} \lor C_{139}
\end{dmath}
\setlength\parindent{20pt}

Similarly, the number of at-least-one clauses for each row, box, and column will reduce. As a reminder, each sub-group clause states that each digit must appear at least once in each sub-group. With the maximum encoding there are nine at-least-one clauses for each sub-group. This number will reduce significantly as the given digits do not need a clause for that sub-group. For instance, the first row of Figure \ref{Medium Sudoku}, has five clauses, skipping the clauses for digits 8, 4, 3, and 1, as they are given digits. The length of those five clauses will reduce based on the number of cells each digit is a candidate in. For example the clause for the digit 5 will be shortened to 1-in-3 SAT clause as the digit 5 only appears as a candidate in three of the five empty cells:
\begin{dmath}
C_{135} \lor C_{155} \lor C_{165}
\end{dmath}
\setlength\parindent{20pt}

This process repeats for all rows, columns, and boxes. Originally, as per the maximum encoding, there were 81 clauses for each of the 4 constraints totaling to 324 1-in-9 SAT clauses. The minimum encoding results in $c$ clauses, where $c$ is the number of empty cells in the Sudoku, for each of the 4 constraints totaling to 4$c$ at-least-one clauses. 

The main reduction in the number of clauses comes in the number of at-most-one clauses. Normally, there are \begin{math}{9 \choose 2} \end{math} clauses per cell, where nine corresponds to the number of literals for any cell, since maximum encoding does not use the candidates. However minimum encoding, using the candidates, reduces the number of at-most-one clauses per cell to \begin{math}{n \choose 2} \end{math} where $n$ represents the number of candidates in any empty cell. For instance, in Figure \ref{Medium Sudoku}, cell (1, 7) now has \begin{math}{3 \choose 2} \end{math} = 3 clauses as supposed to the 36 before:
\begin{dmath}\\
\neg{C_{176}} \lor \neg{C_{177}}\\
\neg{C_{176}} \lor \neg{C_{179}}\\
\neg{C_{177}} \lor \neg{C_{179}}\\
\end{dmath}  
\setlength\parindent{20pt}

This process repeats for the remaining 3 constraints. For each row, column, and box, for any given digit, the number of at-most-one clauses equals \begin{math}{m \choose 2} \end{math} where $m$ represents the number of cells in which that digit appears as a candidate. For example, in box one, or the top left box, of Figure \ref{Medium Sudoku}, the digit 5 appears as a candidate in three cells. The number of  at-most-one clauses for the digit 5 in box one reduces to \begin{math}{3 \choose 2} \end{math} = 3:
\begin{dmath}\\
\neg{C_{135}} \lor \neg{C_{225}}\\
\neg{C_{135}} \lor \neg{C_{235}}\\
\neg{C_{225}} \lor \neg{C_{235}}\\
\end{dmath}  
\setlength\parindent{20pt}

The number of clauses with the use of candidates will reduce based on the number of blanks in the Sudoku. Figure 1 consists of two Sudoku Puzzles from New York Times from two different difficulty levels. Using maximum encoding, the two Sudoku puzzles in Figure \ref{NyTimes Sudoku} are encoded into 12026 and 12011 clauses respectively. With minimum encoding, the two SAT formulations for these two Sudoku puzzles have 545 and 1901 clauses. This is a significant decrease from maximum encoding.

It is worthwhile noting that the at-most-row, at-most-column, and at-most-box clauses are redundant and not required for a SAT solver to solve the Sudoku puzzle. However, these clauses are included for a direct comparison in number of clauses with other sources \cite{chaos} \cite{kwon2006optimized}.

\section{Simulating Human Solvers}\label{simulating human solvers}

The most natural way for early Sudoku practitioners to solve a Sudoku puzzle is through trial-and-error. The adaptation of trial-and-error in Sudoku solving is known as \textit{Nishio} \cite{Eppstein_Nishio}. In Nishio, a candidate value is temporarily placed in a cell, and the Sudoku puzzle is checked for contradictions (an empty cell has no candidates remaining). If a contradiction arises, the candidate value is removed, and a different candidate value is placed in the cell. This process continues by replacing candidate values until the puzzle is either solved or a contradiction is found. If neither happens, a another guess is needed. A new candidate is placed in a different cell, and the process continues.

While this method should work ultimately, it is time-consuming and not practical for human solvers. Instead, by interleaving the Nishio trials with human Sudoku-solving strategies, the solving process can be sped up. More specifically, after guessing a candidate in a cell, these human strategies can be applied to eliminate other candidates and fill cells more efficiently, working toward a quicker solution.

This paper focuses on four Sudoku-solving strategies, ranging from beginner-friendly to moderately sophisticated. These strategies are \textit{Naked Singles}, \textit{Hidden Singles}, \textit{Naked Twins}, and \textit{X-wing}. Each strategy is explained in detail in the following section.

\subsection{Human Strategies} \label{strategies}
\paragraph{Naked Singles}
Naked Singles \cite{solving_sudoku_by_hand} is a natural outcome of the candidates construct described in Section 2. Any empty cells that have only one candidate can be filled with that candidate. For instance, in Figure \ref{Easy Sudoku}, the cell at (1, 2) has just one candidate, the digit 6. Therefore, 6 must be placed in the cell (1, 2). Afterward, the digit 6 can be removed as a candidate from the row, column, and box which cell (1, 2) is in.

\paragraph{Hidden Singles}
Hidden Singles \cite{solving_sudoku_by_hand}  is very similar to Naked Singles but a little more challenging to spot, as suggested by its name. A Hidden Single occurs when a number can only be placed in one specific cell within a row, column, or 3x3 box, even if other candidates have been identified for that cell. To find a Hidden Single, each row, column, and 3x3 box must be scanned to determine whether a digit appears as a candidate in only one empty cell. If such a digit exists, then it must be placed in that cell. 
For example, in row 6 of Figure \ref{Easy Sudoku}, the digit 5 only appears in the first cell (6, 1) as a candidate and in no other cell in that row. Therefore, 5 must be placed in cell (6, 1). Once placed, 5 can be removed as a candidate from the intersecting column and box, narrowing down possibilities for the remaining cells. In Figure \ref{Easy Sudoku}, the digit 5 can be removed as a candidate from the following cells: (7, 1) and (9, 1).

\paragraph{Naked Twins}
Naked Twins \cite{solving_sudoku_by_hand}, also known as Naked Pairs, is an intermediate level Sudoku strategy that is slightly more sophisticated than Naked Singles and Hidden Singles. This strategy arises when two cells in a row, column, or 3x3 box share the exact same pair of candidates and do not contain any other candidates. Since these two candidates must occupy those two cells, they can be eliminated from the candidate list of other cells in the same row, column, or box.
For example, in column 3 of Figure \ref{Easy Sudoku}, the fourth and fifth cells share the same candidates: 2 and 8. As a result, these candidates can be removed from the list of possible candidates for the other cells in the row, column, and box intersecting these two cells. In Figure \ref{Easy Sudoku}, the digit 2 can be  removed from cells (2, 3), (4, 2), (5, 1), (5, 5), (6, 1), and (8, 3), while the digit 8 can be removed from cells (5, 6) and (5, 9).

\begin{figure}[H]
    \centering
    \rule[1ex]{\textwidth}{0.1pt}
    \includegraphics[width=0.5\linewidth]{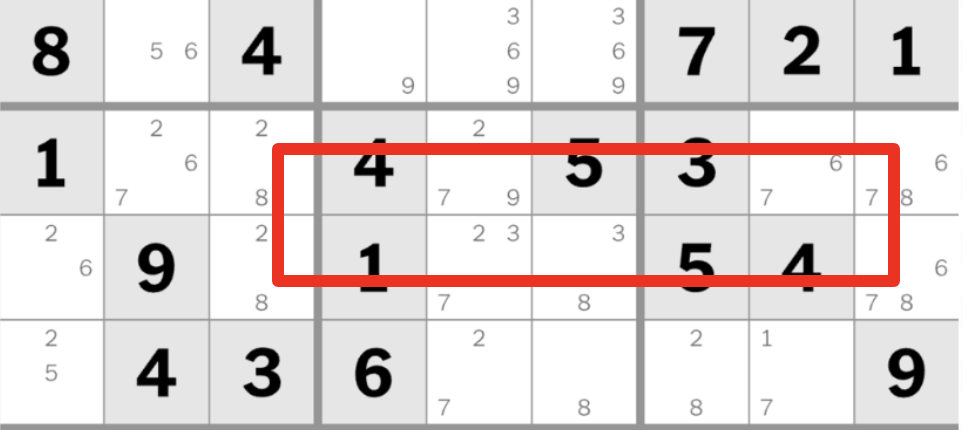}
    \caption{A rectangle is drawn between the cells where X-wing should be used}
    \label{SudokuXWing}
    \rule[1ex]{\textwidth}{0.1pt}
\end{figure}

\paragraph{X-wing}
X-wing is an advanced, yet relatively beginner-friendly strategy that can greatly help in solving complex Sudoku puzzles. X-wing is identified by finding two rows (or columns) where a specific candidate appears exactly twice. Then, check if these occurrences align within the same columns (or rows) to form a rectangle shape. If this pattern exists, that candidate can be removed from other cells within those same columns (or rows).
For instance, in Figure \ref{Easy Sudoku}, the 8 appears only in rows 4 and 5 of columns 3 and 9, creating a rectangle pattern as seen in Figure \ref{SudokuXWing}. This configuration makes it possible to apply the X-wing strategy by removing 8 from all other cells in rows 4 and 5. A more detailed explanation for X-wing is provided by Brouwer \cite{X-wing}. While this example only removes 8 as a candidate from one cell, (5, 6), X-wing can often eliminate many more candidates in more difficult puzzles.

\subsection{Simulating Human Solver via Randomized Nishio with Human Strategies}

Armed with the above knowledge, I now state the assumptions on how a human solver, knowing these strategies, will solve Sudoku puzzles. I have made these assumptions loosely based on how I, a beginner to Sudoku, solved Sudoku puzzles. Going forward, human solvers will be referred to as Hsolvers. 

\begin{enumerate}
    \item The Hsolver prefers simple strategies over difficult strategies. Therefore, strategies are applied in order of increasing difficulty.  The order in which the strategies are applied is Naked Singles, Hidden Singles, Naked Twins, and X-wing. 
    \item Hsolvers know at least 2 strategies and at most 4 strategies.
        \begin{itemize}
            \item If the Hsolver uses 2 strategies, they are Naked Singles and Hidden Singles.
            \item If the Hsolver uses 3 strategies, they are Naked Singles, Hidden Singles, and Naked Twins.
            \item If the Hsolver uses 4 strategies, they are Naked Singles, Hidden Singles, Naked Twins, and X-wing.
        \end{itemize}
    \item When performing Nishio, a \textit{random} cell is chosen each time to assume a candidate, and a \textit{random} candidate is assumed. I assume this for the simulation since using one optimized methodology could lead to idiosyncratic results. The choosing of the cell and candidate is randomized to account for Hsolvers of different skill levels and enable proper characterization of Sudoku difficulty across websites.
\end{enumerate}

  Based on these assumptions, Figure \ref{randomized nishio} describes the steps for the Nishio interleaved with Human Strategies method used in the simulation of human solvers.  
\begin{figure}[H]
    \centering
    \rule[1ex]{\textwidth}{0.1pt}
    \begin{enumerate}
    \item Candidates are computed for every cell (as described in Section \ref{Running Example and Candidates}). 
    \item Start of Nishio; a \textit{random} cell is chosen to assume a candidate, and a \textit{random} candidate is assumed in that cell,
    \item Using the candidate assumed in step 2, the four human strategies are applied repeatedly until the puzzle cannot be progressed anymore. Note that each strategy is applied repeatedly until it cannot make progress before applying the next strategy (order described above). 
    \item When puzzle progress halts, there are exactly three possible actions: 
        \begin{itemize}
            \item The puzzle is solved and no action is needed.
            \item There is at least one cell with no candidates left. In this case eliminate the candidate assumed in step 3 as a candidate and return to step 3.  
            \item The puzzle is not solved and there are no cells with no candidates, meaning another guess is needed. In this case retain the current puzzle state and return to step 3.
        \end{itemize}
    \end{enumerate}
    \caption{Randomized Nishio Interleaving Human Strategies}
    \label{randomized nishio}
    \rule[1ex]{\textwidth}{0.1pt}
\end{figure}

\section{Datasets Used in this Study}
For this study, puzzles from five popular Sudoku websites have been collected, along with the difficulty level of each puzzle in accordance with that of the website. Most websites have an archive, including Sudoku.org.uk, Extreme Sudoku, and Sudoku of the Day UK. Some websites have limited archives such as Sudoku of the Day, and others, such as New York Times, have no archive. For these websites, I have collected puzzles for each difficulty over an extended period of time to ensure that every difficulty level in each website has an equal number of puzzles. The dataset consists of a total of 1320 puzzles, 60 from each difficulty level in each website. \footnote[1]{The permission to publicly release the datasets used in this study from Sudoku.org.uk, Extreme Sudoku, and Sudoku of the Day has been obtained. The datasets are available for academic purposes at https://github.com/ArmanE-V/Sudoku-Difficulty.}

\paragraph{New York Times} New York Times \cite{NYT} releases three puzzles daily, each of a different difficulty level - Easy, Medium, and Hard. This subset contains a total of 180 puzzles collected between March 11, 2024, and May 9, 2024. To the best of my knowledge, a New York Times Sudoku archive does not exist. 

\paragraph{Sudoku.org.uk} Sudoku.org.uk \cite{Sudoku.org.uk} releases one puzzle everyday, with each day corresponding to one of four difficulty levels, with the release schedule as follows: Gentle - Monday, Moderate - Tuesday/Wednesday/Saturday, Tough - Thursday, Diabolical - Friday/Sunday.This subset contains a total of 240 puzzles collected between April 15, 2024, and June 5, 2025. Sudoku.org.uk has a publicly available archive for all its released puzzles. 

\paragraph{Extreme Sudoku} Extreme Sudoku \cite{Extreme_Sudoku} releases five puzzles daily, each of a different difficulty level - Evil, Excessive, Egregious, Excruciating, and Extreme. This subset contains a total of 300 puzzles collected between December 5, 2024, and February 2, 2025. Extreme Sudoku has a publicly available archive for all its released puzzles. 

\paragraph{Sudoku of the Day} Sudoku of the Day \cite{SOFTD} releases six puzzles daily, each of a different difficulty level - Beginner, Easy, Medium, Tricky,  Fiendish, and Diabolical. This subset contains a total of 360 puzzles collected between December 17, 2024, and February 20, 2025. Sudoku of the day publishes an archive for the previous week of released Sudokus. 

\paragraph{Sudoku of the Day UK} Sudoku of the Day UK \cite{SOFTDUK} releases four puzzles daily, each of a different difficulty level - Easy, Medium, Hard, and Absurd. Within each level, the website has a star rating for each puzzle, ranging from 0 to 5 stars, to represent the degree of difficulty. A 0 star Absurd puzzle is harder than a 5 star Hard puzzle. This subset contains a total of 240 puzzles collected between December 1, 2024, and January 29, 2025. Sudoku of the Day UK has a publicly available archive for all its released puzzles.


\section{Results}\label{results}
The results are divided into two sections, one for each method,  encoding Sudoku puzzles as SAT problems and Randomized Nishio Interleaving Human Strategies. Within each section are described the metric measured for that specific method. For the SAT method, I compute Clause Length Distribution. For Randomized Nishio with Human Strategies, Nishio Human Cycles are computed, measuring the logical steps taken by the Hsolver in the solving process. Finally, by applying only the human strategies, I compute the percentage of puzzles solved by human strategies alone in each difficulty level. It is important to note that only Nishio Human Cycles and Clause Length Distribution are proposed metrics. Percentage of puzzles solved by human strategies alone is an additional dimension for which I preform analysis on and provide results. 

All experiments were run on an Apple MacBook Air 2020 with an Apple M1 chip and 8 GB RAM. I use the Glucose3\cite{Glucose3} SAT solver for all SAT experiments. 
\subsection{SAT}\label{SAT results}

\setlength\parindent{20pt} The mean Clause Length Distribution for each difficulty level is displayed in Table \ref{SAT Table}

\begin{enumerate}
\item \textbf{Clause Length Distribution}\footnote[2]{Note that all at-most-one clauses are binary and will skew the data towards $k = 2$, and therefore only at-least-one clauses are considered in the Clause Length Distribution.} Each at-least-one clause, using the encoding described in Section \ref{candidate encoding}, is a 1-in-$k$ SAT clause. $k$ represents the length in number of literals in that clause and each clause can have length one (unit clause) to length nine (9-ary clause). Let $T$ equal the total number of at-least-one clauses and $f_k$ represent the frequency of clauses with length $k$. The percentage of at-least-one clauses with length $k$ is \begin{math}P_k  = \frac{f_k}{T} * 100\end{math}. Clause Length Distribution displays the distribution of percentages across clauses lengths 1-9. Further, Clause Length Distribution categorizes clause lengths into three categories. Clauses with $k = 1, 2$, are 'short' clauses, clauses with $k = 3, 4, 5$ are 'medium' clauses, and k= $6, 7, 8, 9$  are 'long' clauses. For example, in Table \ref{SAT Table}, the mean percentage of short clauses for New York Times Easy is 29\%,i.e., on average, across 60 New York Times Easy Sudoku Puzzles, 29\% of the at-least-one clauses are short.
\end{enumerate}

\begin{table}[H]
\caption{Clause Length Distribution across 5 websites}
\resizebox{\textwidth}{!}{
\begin{tabular}{clclcllclll}
\multicolumn{2}{c}{Website}                             & \multicolumn{2}{c}{Short Clauses (\%)} & \multicolumn{3}{c}{Medium Clauses (\%)} & \multicolumn{4}{c}{Long Clauses (\%)} \\ \hline
\multicolumn{1}{l}{New York Times}       &              &                    &                   &             &             &             &         &         &         &         \\
                                         & Easy         & \multicolumn{2}{c}{67}                 & \multicolumn{3}{c}{33}                  & \multicolumn{4}{c}{0}                 \\
                                         & Medium       & \multicolumn{2}{c}{17}                 & \multicolumn{3}{c}{71}                  & \multicolumn{4}{c}{13}                \\
                                         & Hard         & \multicolumn{2}{c}{15}                 & \multicolumn{3}{c}{72}                  & \multicolumn{4}{c}{13}                \\
\multicolumn{1}{l}{Sudoku.org.uk}        &              &                    &                   &             &             &             &         &         &         &         \\
                                         & Gentle       & \multicolumn{2}{c}{24}                 & \multicolumn{3}{c}{67}                  & \multicolumn{4}{c}{9}                 \\
                                         & Moderate     & \multicolumn{2}{c}{21}                 & \multicolumn{3}{c}{69}                  & \multicolumn{4}{c}{10}                \\
                                         & Tough        & \multicolumn{2}{c}{19}                 & \multicolumn{3}{c}{71}                  & \multicolumn{4}{c}{10}                \\
                                         & Diabolical   & \multicolumn{2}{c}{19}                 & \multicolumn{3}{c}{71}                  & \multicolumn{4}{c}{10}                \\
\multicolumn{1}{l}{Extreme Sudoku}       &              &                    &                   &             &             &             &         &         &         &         \\
                                         & Evil         & \multicolumn{2}{c}{20}                 & \multicolumn{3}{c}{72}                  & \multicolumn{4}{c}{7}                 \\
                                         & Excessive    & \multicolumn{2}{c}{20}                 & \multicolumn{3}{c}{73}                  & \multicolumn{4}{c}{7}                 \\
                                         & Egregious    & \multicolumn{2}{c}{20}                 & \multicolumn{3}{c}{73}                  & \multicolumn{4}{c}{6}                 \\
                                         & Excruciating & \multicolumn{2}{c}{20}                 & \multicolumn{3}{c}{72}                  & \multicolumn{4}{c}{7}                 \\
                                         & Extreme      & \multicolumn{2}{c}{20}                 & \multicolumn{3}{c}{73}                  & \multicolumn{4}{c}{7}                 \\
\multicolumn{1}{l}{Sudoku of the Day}    &              &                    &                   &             &             &             &         &         &         &         \\
                                         & Beginner     & \multicolumn{2}{c}{56}                 & \multicolumn{3}{c}{43}                  & \multicolumn{4}{c}{1}                 \\
                                         & Easy         & \multicolumn{2}{c}{37}                 & \multicolumn{3}{c}{59}                  & \multicolumn{4}{c}{4}                 \\
                                         & Medium       & \multicolumn{2}{c}{23}                 & \multicolumn{3}{c}{69}                  & \multicolumn{4}{c}{8}                 \\
                                         & Tricky       & \multicolumn{2}{c}{21}                 & \multicolumn{3}{c}{71}                  & \multicolumn{4}{c}{8}                 \\
                                         & Fiendish     & \multicolumn{2}{c}{19}                 & \multicolumn{3}{c}{72}                  & \multicolumn{4}{c}{9}                 \\
                                         & Diabolical   & \multicolumn{2}{c}{20}                 & \multicolumn{3}{c}{71}                  & \multicolumn{4}{c}{9}                 \\
\multicolumn{1}{l}{Sudoku of the Day UK} &              &                    &                   &             &             &             &         &         &         &         \\
                                         & Easy         & \multicolumn{2}{c}{24}                 & \multicolumn{3}{c}{66}                  & \multicolumn{4}{c}{10}                \\
                                         & Medium       & \multicolumn{2}{c}{17}                 & \multicolumn{3}{c}{70}                  & \multicolumn{4}{c}{13}                \\
                                         & Hard         & \multicolumn{2}{c}{16}                 & \multicolumn{3}{c}{70}                  & \multicolumn{4}{c}{14}                \\
                                         & Absurd       & \multicolumn{2}{c}{17}                 & \multicolumn{3}{c}{72}                  & \multicolumn{4}{c}{11}                \\ \hline
\end{tabular}
}
\label{SAT Table}
\end{table}

\subsection{Nishio Human Cycles}\label{Nishio Results}
Each puzzle has been simulated using 50 different random starting points (the random order in which cells are chosen to assume a candidate and the random order of which candidate will be assumed in each cell; see step 2 in Figure \ref{randomized nishio}) using the solver in Figure \ref{randomized nishio}. A metric measured in this process is "Nishio Human Cycles". The application of all four human strategies once in step 3 in Figure \ref{randomized nishio} is counted as one cycle. The solver repeats these cycles until they no longer affect the puzzle or candidates, at which point the solver proceeds to step 4 in Figure \ref{randomized nishio}. The solver counts the total number of cycles performed in solving the Sudoku puzzle. The Nishio Human Cycles value for one Sudoku puzzle is the mean number of cycles needed to solve the same Sudoku puzzle over 50 solves. Solving the puzzle 50 times allows us to simulate 50 different Hsolvers solving the same Sudoku puzzle.  Please note that since the choosing of the cell and candidate is randomized (i.e. the 50 random starting points) for every solve, each solve will not require the same number of cycles. Additionally, I run the solver in Figure \ref{randomized nishio} on the 1320 puzzle dataset, replacing 4 strategies with only 2 strategies (Naked Singles and Hidden Singles) in step 3. The results for Nishio Human Cycles for 2 strategies, represents the total number of cycles using only 2 strategies performed in solving the puzzle. Note that the 50 random starting points used for Nishio Human Cycles with 4 strategies are not the same as used for Nishio Human Cycles with 2 strategies. 

\subsection{Percentage of Puzzles Solved by Human Strategies Alone} Without applying Nishio, and just using strategies, it is possible to solve some Sudoku puzzles. The strategies are used repeatedly until either the puzzle is solved or no progress is possible. Percentage of Puzzles Solved by Human Strategies Alone measures the percentage of puzzles solved by using 2 , 3 , and 4 strategies. In Table \ref{SAT Table}, for the column labeled 2 strategies, Naked Singles and Hidden Singles are used, in that order. Similarly for 3 strategies, Naked Singles, Hidden Singles, and Naked Twins are applied, in that order, and finally, for 4 strategies, Naked Singles, Hidden Singles, Naked Twins, and X-wing are applied, in that order. 

Table \ref{NyTimes Table} to Table \ref{Sudoku of the Day UK Table} display the results for percentage of puzzles solved by human strategies alone and Nishio Human Cycles with 2 and 4 strategies for every difficulty level in each individual website. 

\section{Analysis}\label{Analysis}
Analysis of the results is divided into two parts: within site and cross-site.

\textbf{Within Site} analysis is performed over three dimensions given below. For each dimension I make assertions that capture the intuition behind the that dimension. 
\begin{enumerate}
    \item Clause Length Distribution Assertions
        \begin{itemize}
            \item As puzzle difficulty increases, the percentage of short clauses should decrease monotonically.
            \item As puzzle difficulty increases, the percentage of medium and long clauses should increase monotonically.
        \end{itemize}     
    \item Percentage of Puzzles solved by only Human Strategies Assertions
    \begin{itemize}
        \item For a fixed number of human strategies (2, 3, 4), the  percentage of puzzles solved should increase monotonically as difficulty level increases. 
        \item For puzzles in a fixed difficulty level, increasing the number of human strategies from 2 to 3 to 4 should increase monotonically the percentage of puzzles solved. 
    \end{itemize}
    \item Nishio Human Cycles Assertions
        \begin{itemize}
            \item The average number of Nishio Human Cycles required to solve a puzzle should increase monotonically with difficulty.
            \item Within the same difficulty level, the difference in  Nishio Human Cycles between using 2 human strategies versus 4 strategies should increase monotonically as difficulty level increases. 
        \end{itemize}  
\end{enumerate}

\textbf{Cross-Site} analysis establishes similarities and differences across the Sudoku websites. To compare difficulty levels across the 5 Sudoku websites, I introduce a universal classification system based on an unsupervised algorithm (described in Section \ref{universal classification}) that partitions puzzles into three bins using the distributions of Nishio Human Cycles and percentage of short clauses. This universal classification allows us to compare the five website’s difficulty levels highlighting key patterns, consistencies, and outliers observed across these websites.

\subsection{Within Site Analysis} \label{within site analysis}
\subsubsection{Spearman's Rank Correlation Coefficient for all Websites} \label{spearman's section}
\begin{table}[H]
\centering
\captionsetup{width=.75\textwidth}
\caption{\textit{\textbf{Spearman’s Rank Correlation Coefficient between website-provided difficulty levels and two proposed metrics - Clause Length Distribution and Nishio Human Cycles. For Clause Length Distribution, I use the percentage of short clauses as the metric to calculate the correlation coefficient. Each puzzle is assigned an integer ranking based on the website's difficulty level ordering (e.g. in New York Times, Easy puzzles are assigned 1, Medium puzzles are assigned 2, and Hard puzzles are assigned 3).}}}
\begin{tabular}{lllccc}
\hline
Website              &  &  & \begin{tabular}[c]{@{}c@{}}Clause Length\\ Distribution\end{tabular} & \begin{tabular}[c]{@{}c@{}}Nishio Human Cycles with \\ 4 Human Strategies\end{tabular} & \multicolumn{1}{l}{\begin{tabular}[c]{@{}l@{}}Nishio Human Cycles with \\ 2 Human Strategies\end{tabular}} \\ \hline
NyTimes Sudokus      &  &  & -0.80                                                                & 0.84                                                                            & 0.68                                                                                                \\
Sudoku.org.uk        &  &  & -0.47                                                                & 0.84                                                                            & 0.81                                                                                                \\
Extreme Sudoku       &  &  & -0.02                                                                & -0.002                                                                          & -0.03                                                                                               \\
Sudoku of the Day    &  &  & -0.75                                                                & 0.89                                                                            & 0.83                                                                                                \\
Sudoku of the Day UK &  &  & -0.28                                                                & 0.70                                                                            & 0.61                                                                                                \\ \hline
\end{tabular}
\label{Spearman's}
\end{table}
Table \ref{Spearman's} shows the Spearman's rank correlation coefficient between the website label difficulties and the proposed metrics. Nishio Human Cycles using four strategies has higher correlation with website labels than Nishio Human Cycles with 2 strategies. As a result, subsequent analysis will focus mainly on Nishio Human Cycles with 4 strategies. Clause length distribution shows high inverse correlation (expected) for New York Times and Sudoku of the Day. For Sudoku.org.uk and Sudoku of the Day UK, the metric shows moderate correlation. Extreme Sudoku has a near-zero correlation coefficient for both proposed metrics. Also, notice that Nishio Human Cycles with 4 Human Strategies has a higher correlation with website-labeled difficulty than Clause Length Distribution for 4 out of 5 websites.

Before proceeding further, it is important to note the conceptual difference between the two metrics used in this study. Clause length distribution captures the structural complexity of a puzzle, including the number of given digits, the specific values used as given digits, and the cells they are in. In contrast, Nishio Human Cycles captures the procedural difficulty in solving the puzzle. 

\begin{table}[H]
\caption{New York Times}
\resizebox{\textwidth}{!}{
\begin{tabular}{clcclccc}
\multicolumn{1}{l}{}  &        & \multicolumn{2}{c}{Nishio Human Cycles (50 Random Starts)}                        &  & \multicolumn{3}{c}{Percentage of Puzzles solved by Strategies}       \\ \cline{3-4} \cline{6-8} 
\multicolumn{1}{l}{}  &        & \multicolumn{1}{l}{}             & \multicolumn{1}{l}{} &  & \multicolumn{1}{l}{} & \multicolumn{1}{l}{}  & \multicolumn{1}{l}{}  \\
\multicolumn{2}{c}{Difficulty} & \multicolumn{1}{l}{4 Strategies} & 2 Strategies         &  & 2 Strategies (\%)    & 3 Strategies (\%)     & 4 Strategies (\%)     \\ \hline
\multicolumn{8}{l}{Easy}                                                                                                                                      \\
                      & Mean   & 1.61                             & 1.62                 &  & \multirow{2}{*}{100} & \multirow{2}{*}{100}  & \multirow{2}{*}{100}  \\
                      & Median & 1.58                             & 1.60                &  &                      &                       &                       \\
\multicolumn{8}{l}{Medium}                                                                                                                                    \\
                      & Mean   & 6.74                             & 12.40                &  & \multirow{2}{*}{0}   & \multirow{2}{*}{52.5} & \multirow{2}{*}{62.5} \\
                      & Median & 5.32                             & 9.98                 &  &                      &                       &                       \\
\multicolumn{8}{l}{Hard}                                                                                                                                      \\
                      & Mean   & 13.18                            & 29.92                &  & \multirow{2}{*}{0}   & \multirow{2}{*}{24}   & \multirow{2}{*}{32.5} \\
                      & Median & 10.03                            & 21.55                &  &                      &                       &                       \\ \hline
\end{tabular}
}
\label{NyTimes Table}
\end{table}
\subsubsection{New York Times}
\begin{enumerate}

    \item \textbf{Clause Length Distribution} \\
    All assertions described in Section \ref{Analysis} hold true, with Medium and Hard have similar Clause Length Distributions.  
    \item \textbf{Percentage of Puzzles solved by only Human Strategies} \\
    All assertions hold true. All Easy puzzles can be solved using just 2 strategies and the Hsolver must know at least 3 strategies to solve any Medium or Hard puzzles. 
    \item  \textbf{Nishio Human Cycles}\\
     All assertions hold true. There are substantial differences in mean and median Nishio Human Cycles between the three difficulty levels indicating sizeable increases in complexity when moving from Easy to Medium to Hard.
\end{enumerate}
\begin{table}[H]
\caption{Sudoku.org.uk}
\resizebox{\textwidth}{!}{
\begin{tabular}{clcclccc}
\multicolumn{1}{l}{}  &        & \multicolumn{2}{c}{Nishio Human Cycles (50 Random Starts)}                        &  & \multicolumn{3}{c}{Percentage of Puzzles solved by Strategies}     \\ \cline{3-4} \cline{6-8} 
\multicolumn{1}{l}{}  &        & \multicolumn{1}{l}{}             & \multicolumn{1}{l}{} &  & \multicolumn{1}{l}{} & \multicolumn{1}{l}{} & \multicolumn{1}{l}{} \\
\multicolumn{2}{c}{Difficulty} & \multicolumn{1}{l}{4 Strategies} & 2 Strategies         &  & 2 Strategies (\%)    & 3 Strategies (\%)    & 4 Strategies (\%)    \\ \hline
\multicolumn{8}{l}{Gentle}                                                                                                                                  \\
                      & Mean   & 2.45                             & 2.51                 &  & \multirow{2}{*}{98} & \multirow{2}{*}{98} & \multirow{2}{*}{98} \\
                      & Median & 2.38                             & 2.4                 &  &                      &                      &                      \\
\multicolumn{8}{l}{Moderate}                                                                                                                                \\
                      & Mean   & 3.09                             & 3.77                 &  & \multirow{2}{*}{97}  & \multirow{2}{*}{98}  & \multirow{2}{*}{98}  \\
                      & Median & 2.96                             & 3.04                  &  &                      &                      &                      \\
\multicolumn{8}{l}{Tough}                                                                                                                                   \\
                      & Mean   & 5.93                             & 9.72                 &  & \multirow{2}{*}{0}   & \multirow{2}{*}{38}  & \multirow{2}{*}{53}  \\
                      & Median & 4.61                             & 8.31                  &  &                      &                      &                      \\
\multicolumn{8}{l}{Diabolical}                                                                                                                              \\
                      & Mean   & 7.94                             & 10.41                &  & \multirow{2}{*}{1.5}   & \multirow{2}{*}{3}   & \multirow{2}{*}{3}   \\
                      & Median & 7.04                              & 9.35                  &  &                      &                      &                      \\ \hline
\end{tabular}
}
\label{Sudoku.org.uk Table}
\end{table}
\subsubsection{Sudoku.org.uk}
\begin{enumerate}
    \item \textbf{Clause Length Distribution}\\
    All assertions hold true. Note that Tough and Diabolical have identical Clause Length Distributions.
    \item \textbf{Percentage of Puzzles solved by only Human Strategies}\\
    There is a slight increase in percentage of puzzles solved by 2 strategies from Tough to Diabolical,  but the increase is negligible. Diabolical has a decidedly smaller percentage of puzzles solved by 3 and 4 strategies, consistent with the assertion. All other assertions hold true. 
    \item  \textbf{Nishio Human Cycles}\\
    All assertions hold true. 
    \end{enumerate}

\begin{table}[H]
\caption{Extreme Sudoku}
\resizebox{\textwidth}{!}{
\begin{tabular}{clcclccc}
\multicolumn{1}{l}{}  &        & \multicolumn{2}{c}{Nishio Human Cycles (50 Random Starts)}                        &                      & \multicolumn{3}{c}{Percentage of Puzzles solved by Strategies}     \\ \cline{3-4} \cline{6-8} 
\multicolumn{1}{l}{}  &        & \multicolumn{1}{l}{}             & \multicolumn{1}{l}{} &                      & \multicolumn{1}{l}{} & \multicolumn{1}{l}{} & \multicolumn{1}{l}{} \\
\multicolumn{2}{c}{Difficulty} & \multicolumn{1}{l}{4 Strategies} & 2 Strategies         &                      & 2 Strategies (\%)    & 3 Strategies (\%)    & 4 Strategies (\%)    \\ \hline
\multicolumn{8}{l}{Evil}                                                                                                                                                        \\
                      & Mean   & 8.01                             & 14.01                &                      & \multirow{2}{*}{0}   & \multirow{2}{*}{0}   & \multirow{2}{*}{5}  \\
                      & Median & 6.63                             & 10.09                &                      &                      &                      &                      \\
\multicolumn{8}{l}{Excessive}                                                                                                                                                   \\
                      & Mean   & 7.13                             & 10.89                &                      & \multirow{2}{*}{0}   & \multirow{2}{*}{0}   & \multirow{2}{*}{3}   \\
                      & Median & 5.98                             & 9.40                &                      &                      &                      &                      \\
\multicolumn{8}{l}{Egregious}                                                                                                                                                   \\
                      & Mean   & 7.51                             & 11.98                 &                      & \multirow{2}{*}{0}   & \multirow{2}{*}{0}   & \multirow{2}{*}{0}   \\
                      & Median & 5.48                             & 8.16                 &                      &                      &                      &                      \\
\multicolumn{8}{l}{Excruciating}                                                                                                                                                \\
                      & Mean   & 7.39                             & 10.99                 &                      & \multirow{2}{*}{0}   & \multirow{2}{*}{0}   & \multirow{2}{*}{0}   \\
                      & Median & 6.43                             & 8.71                 &                      &                      &                      &                      \\
\multicolumn{8}{l}{Extreme}                                                                                                                                                     \\
                      & Mean   & 7.63                             & 12.07                &                      & \multirow{2}{*}{0}   & \multirow{2}{*}{0}   & \multirow{2}{*}{0}   \\
\multicolumn{1}{l}{}  & Median & 6.86                             & 10.28                 & \multicolumn{1}{c}{} &                      &                      &                      \\ \hline
\end{tabular}
}
\label{Extreme Sudoku Table}
\end{table}
\subsubsection{Extreme Sudoku}
\label{Extreme Sudoku}
\begin{enumerate}
    \item \textbf{Clause Length Distribution} \\
    All five difficulty levels have near identical Clause Length Distributions, with neither assertion holding true.  
     \item \textbf{Percentage of Puzzles solved by only Human Strategies}\\
    Although both assertions hold true, it is important to note that if the Hsolver knew only these four strategies, they can solve only a small fraction of puzzles from the entire website (in Evil and Excessive).
    \item \textbf{Nishio Human Cycles}\\
    Extreme Sudoku is anomalous with none of the assertions holding true, and seemingly no correlation between Nishio Human Cycles and website labeled difficulty levels (as also seen in Table \ref{Spearman's}). In fact, when ranked by empirical difficulty - measured by average Nishio Human Cycles - the ordering from least to most difficult is: Excessive, Excruciating, Egregious, Extreme, Evil. 
     \end{enumerate}

\begin{table}[H]
\caption{Sudoku of the Day}
\resizebox{\textwidth}{!}{
\begin{tabular}{clcclccc}
\multicolumn{1}{l}{}  &        & \multicolumn{2}{c}{Nishio Human Cycles (50 Random Starts)}                        &                      & \multicolumn{3}{c}{Percentage of Puzzles solved by Strategies}     \\ \cline{3-4} \cline{6-8} 
\multicolumn{1}{l}{}  &        & \multicolumn{1}{l}{}             & \multicolumn{1}{l}{} &                      & \multicolumn{1}{l}{} & \multicolumn{1}{l}{} & \multicolumn{1}{l}{} \\
\multicolumn{2}{c}{Difficulty} & \multicolumn{1}{l}{4 Strategies} & 2 Strategies         &                      & 2 Strategies (\%)    & 3 Strategies (\%)    & 4 Strategies (\%)    \\ \hline
\multicolumn{8}{l}{Beginner}                                                                                                                                                    \\
                      & Mean   & 1.73                             & 1.73                 &                      & \multirow{2}{*}{100} & \multirow{2}{*}{100} & \multirow{2}{*}{100} \\
                      & Median & 1.72                             & 1.70                 &                      &                      &                      &                      \\
\multicolumn{8}{l}{Easy}                                                                                                                                                        \\
                      & Mean   & 2.09                             & 2.14                 &                      & \multirow{2}{*}{100} & \multirow{2}{*}{100} & \multirow{2}{*}{100} \\
                      & Median & 2.01                             & 2.04                 &                      &                      &                      &                      \\
\multicolumn{8}{l}{Medium}                                                                                                                                                      \\
                      & Mean   & 3.42                             & 5.61                 &                      & \multirow{2}{*}{13}  & \multirow{2}{*}{70}  & \multirow{2}{*}{87}  \\
                      & Median & 3.19                             & 5.03                &                      &                      &                      &                      \\
\multicolumn{8}{l}{Tricky}                                                                                                                                                      \\
                      & Mean   & 5.37                             & 8.24                 &                      & \multirow{2}{*}{0}   & \multirow{2}{*}{43}  & \multirow{2}{*}{50}  \\
                      & Median & 4.98                             & 7.42                 &                      &                      &                      &                      \\
\multicolumn{8}{l}{Fiendish}                                                                                                                                                    \\
                      & Mean   & 8.11                             & 12.89                &                      & \multirow{2}{*}{0}   & \multirow{2}{*}{5}   & \multirow{2}{*}{15}  \\
\multicolumn{1}{l}{}  & Median & 7.24                             & 10.08                 & \multicolumn{1}{c}{} &                      &                      &                      \\
\multicolumn{8}{l}{Diabolical}                                                                                                                                                  \\
                      & Mean   & 7.55                             & 9.05                 &                      & \multirow{2}{*}{0}   & \multirow{2}{*}{0}   & \multirow{2}{*}{0}   \\
\multicolumn{1}{l}{}  & Median & 6.73                             & 7.88                 & \multicolumn{1}{c}{} &                      &                      &                      \\ \hline
\end{tabular}
}
\label{Sudoku of the Day Table}
\end{table}
\subsubsection{Sudoku of the Day}
\begin{enumerate}
    \item \textbf{Clause Length Distribution} \\
    Assertions hold true from Beginner to Fiendish. From Fiendish to Diabolical there is a 1 \% increase in percentage of short clauses and a 1 \% decrease in percentage of medium clauses. Nonetheless, this inconsistency is small and the overall expected trend holds true.  
    \item \textbf{Percentage of Puzzles solved by only Human Strategies} \\
    All assertions hold true. Notice that there is a considerable decrease in percentage of puzzles solved with two strategies from Easy to Medium, indicating a need for more advanced strategies starting at Medium. For Diabolical, Nishio is required to solve all puzzles, knowing only these four strategies. 
    \item  \textbf{Nishio Human Cycles} \\
    All assertions hold true apart from in Fiendish and Diabolical where mean and median Nishio Human Cycles decrease from Fiendish to Diabolical, similar to the discrepancy found in Clause Length Distribution. Since Diabolical sees no puzzles solved using only human strategies, it implies that Sudoku of the Day emphasizes the sophistication of known human strategies rather than how often they are applied.
    \end{enumerate}
\begin{table}[H]
\caption{Sudoku of the Day UK}
\resizebox{\textwidth}{!}{
\begin{tabular}{clcclccc}
\multicolumn{1}{l}{}  &        & \multicolumn{2}{c}{Nishio Human Cycles (50 Random Starts)}                        &  & \multicolumn{3}{c}{Percentage of Puzzles solved by Strategies}     \\ \cline{3-4} \cline{6-8} 
\multicolumn{1}{l}{}  &        & \multicolumn{1}{l}{}             & \multicolumn{1}{l}{} &  & \multicolumn{1}{l}{} & \multicolumn{1}{l}{} & \multicolumn{1}{l}{} \\
\multicolumn{2}{c}{Difficulty} & \multicolumn{1}{l}{4 Strategies} & 2 Strategies         &  & 2 Strategies (\%)    & 3 Strategies (\%)    & 4 Strategies (\%)    \\ \hline
\multicolumn{8}{l}{Easy}                                                                                                                                  \\
                      & Mean   & 3.48                             & 5.20                 &  & \multirow{2}{*}{82} & \multirow{2}{*}{93} & \multirow{2}{*}{95} \\
                      & Median & 2.83                             & 3.43                 &  &                      &                      &                      \\
\multicolumn{8}{l}{Medium}                                                                                                                                \\
                      & Mean   & 4.01                             & 5.49                 &  & \multirow{2}{*}{78}  & \multirow{2}{*}{90}  & \multirow{2}{*}{90}  \\
                      & Median & 3.44                             & 3.67                  &  &                      &                      &                      \\
\multicolumn{8}{l}{Hard}                                                                                                                                   \\
                      & Mean   & 7.28                             & 13.62                 &  & \multirow{2}{*}{42}   & \multirow{2}{*}{53}  & \multirow{2}{*}{55}  \\
                      & Median & 5.84                             & 7.74                  &  &                      &                      &                      \\
\multicolumn{8}{l}{Absurd}                                                                                                                              \\
                      & Mean   & 12.90                            & 16.89                &  & \multirow{2}{*}{0}   & \multirow{2}{*}{0}   & \multirow{2}{*}{0}   \\
                      & Median & 11.01                              & 14.25                  &  &                      &                      &                      \\ \hline
\end{tabular}
}
\label{Sudoku of the Day UK Table}
\end{table}
\subsubsection{Sudoku of the Day UK}
    \begin{enumerate}
    \item \textbf{Clause Length Distribution} \\
    All assertions hold true apart from an increase in percentage of short puzzles and a decrease in percentage of long puzzles from Hard to Absurd. Given other discrepancies found in previous sites and the weak correlation between website labeled difficulty and Clause Length Distribution (as seen in Table \ref{Spearman's}), it is safe to neglect such minor anomalies.  
    \item \textbf{Percentage of Puzzles solved by only Human Strategies}\\
    All assertions hold true. Similar to Diabolical in Sudoku of the Day, Absurd has no puzzles solved by only applying human strategies.
   
    \item \textbf{Nishio Human Cycles}\\
    All assertions hold true. 
     \end{enumerate}

\subsection{Cross-Site Analysis}

\subsubsection{Universal Rating System based on a Simple Unsupervised Classifier}\label{universal classification}
I propose a simple, unsupervised classification method that enables classification of Sudoku difficulty across different websites and individual puzzles. This method uses the metrics described in Section \ref{results} and no site specific labels, thereby allowing for a universally applicable rating system. 
\paragraph{Method Overview}
The simple classifier is built with an \textbf{unsupervised univariate binning} algorithm applied to the two proposed difficulty metrics separately, permitting two different rating systems:
\begin{itemize}
    \item \textbf{Nishio Human Cycles}, measured using the simulation that interleaves Nishio with 4 human strategies. 
    \item \textbf{Clause Length Distribution} Since Clause Length Distribution is a distribution and not a single metric value, I use each Sudoku puzzle's percentage of short clauses as the metric (see Section \ref{SAT results}). From now on, when referring to Clause Length Distribution as a metric, I specifically mean the percentage of short clauses. 
\end{itemize}
For each metric, I use the full set of 1320 data points as a single unlabeled dataset and use the unsupervised univariate binning algorithm to identify a range for each bin. 
\paragraph{Binning Design Choices} Two design choices are required for this method: the number of bins and the strategy (equal-bin-width vs. equal-bin-count) for determining the bin ranges.
\begin{enumerate}
    \item \textbf{Number of Bins} I use three bins, consistent with the fewest number of difficulty levels used by any of the websites in the puzzle dataset (New York Times). The three bins is also motivated by the work by \cite{chaos}. More details are provided in Section 10. These bins are assigned the following category names:
    \begin{itemize}
        \item Universal Easy
        \item Universal Medium
        \item Universal Hard
    \end{itemize}
    \item \textbf{Binning Strategy}
    I choose equal-bin-count (each bin contains approximately the same number of values) over equal-bin-width (each bin is of the same width). Equal-bin-count is preferred for distributions of medium or great skewness \cite{binning}. As seen in Figure \ref{Nishio Human Cycles Distribution} and \ref{Clause Length Distribution Histogram}, the distribution of the 1320-puzzle dataset using the two metrics is skewed right.

\begin{figure}[H]
    \centering
    \rule[1ex]{\textwidth}{0.1pt}
    \makebox[\textwidth][c]{\includegraphics[width=1.15\textwidth]{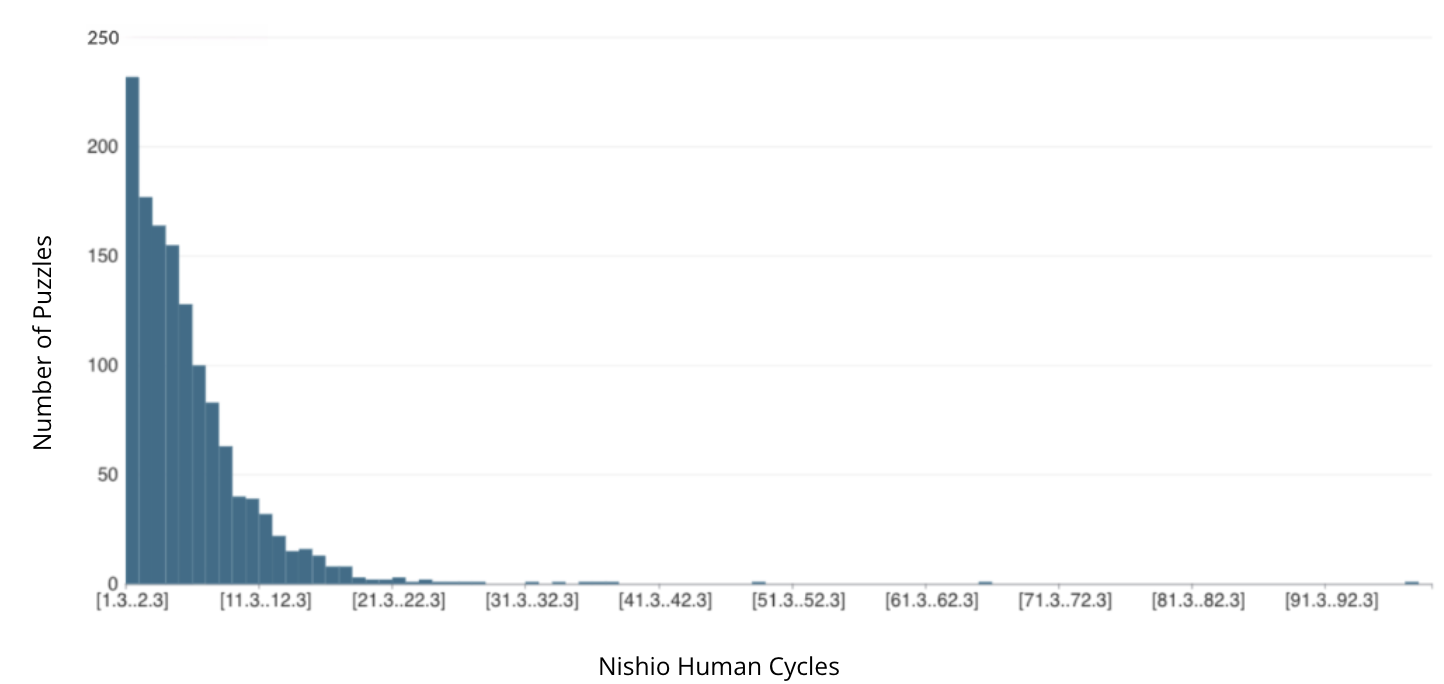}}%
    \caption{Nishio Human Cycles distribution over the 1320 Sudoku puzzle dataset. In this histogram, the width of each bin is is one Nishio Human Cycle.}
    \label{Nishio Human Cycles Distribution}
    \rule[1ex]{\textwidth}{0.1pt}
\end{figure}

\begin{figure}[H]
    \centering
    \rule[1ex]{\textwidth}{0.1pt}
    \makebox[\textwidth][c]{\includegraphics[width=1.15\textwidth]{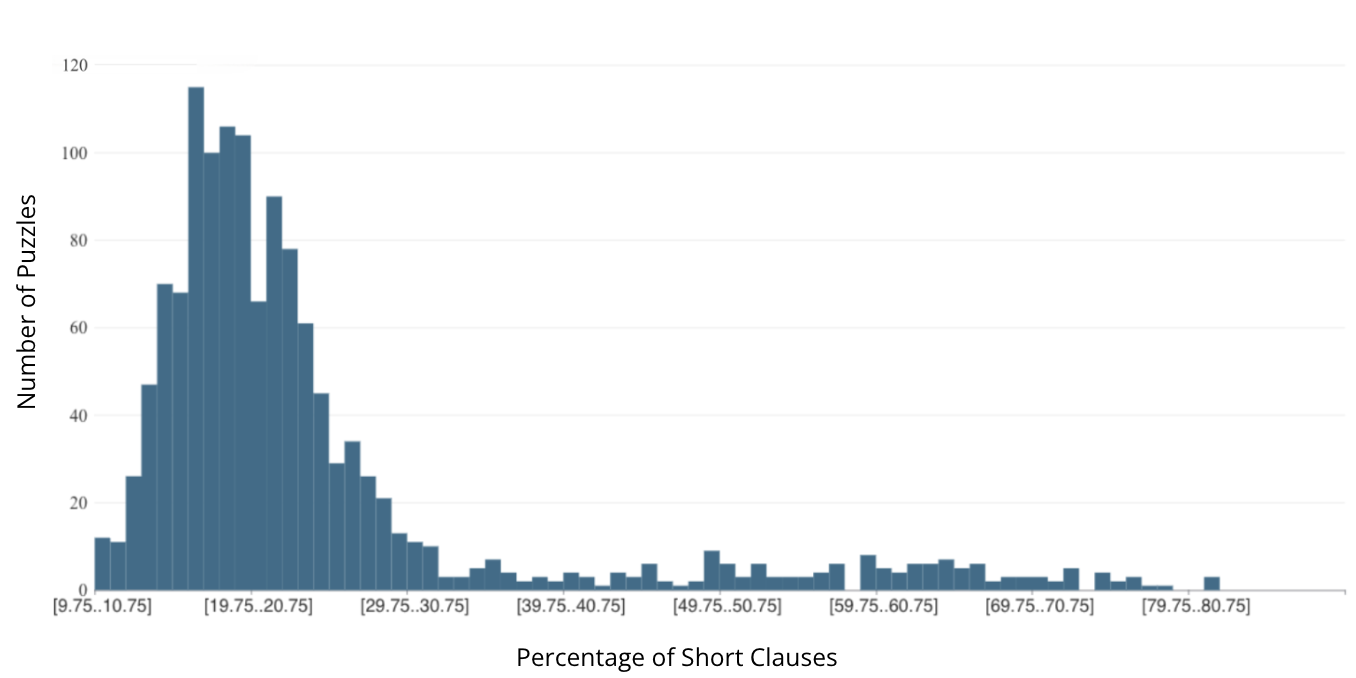}}%
    \caption{Clause Length Distribution over the 1320 Sudoku puzzle dataset. In this histogram, the width of each bin is one percentage (short clauses).}
    \label{Clause Length Distribution Histogram}
    \rule[1ex]{\textwidth}{0.1pt}
\end{figure}

\end{enumerate}
\subsubsection{Mapping Website Difficulty Levels onto Universal Classification}
\paragraph{Mapping Difficulty Levels via Nishio Human Cycles}
Applying univariate binning to the Nishio Human Cycles metric for all 1320 puzzles, returns the following category ranges:
\begin{itemize}
    \item Universal Easy: [1.30, 3.48)
    \item Universal Medium: [3.48, 6.52)
    \item Universal Hard: [6.52, 98.14]
\end{itemize}

These ranges can be applied in two ways:
\begin{enumerate}
    \item To classify individual puzzles directly based on their Nishio Human Cycles value.
    \item To classify entire difficulty levels based on the mean or median Nishio Human Cycles of puzzles in that level.
\end{enumerate}

\begin{figure}[H]
    \centering
    \rule[1ex]{\textwidth}{0.1pt}
    \makebox[\textwidth][c]{\includegraphics[width=1.22\textwidth]{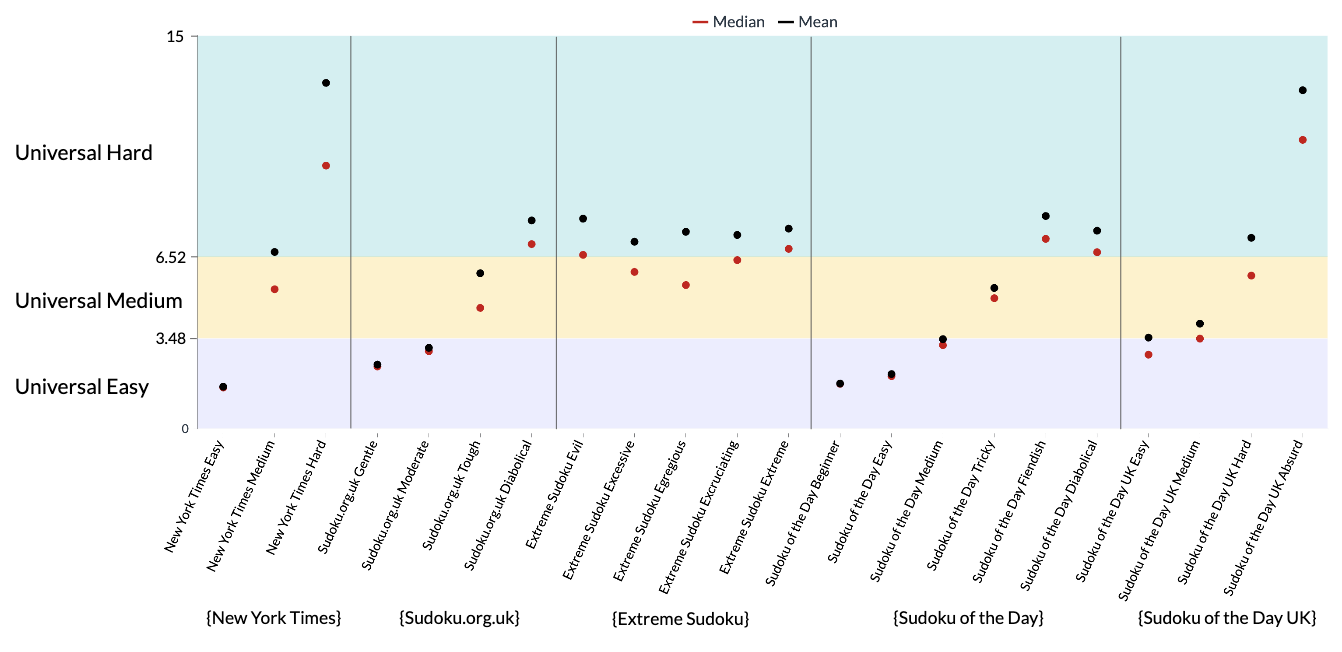}}%
    \caption{Difficulty levels from the 5 websites classified into one of three universal categories based Mean and Median Nishio Human Cycles (from Tables \ref{NyTimes Table} - \ref{Sudoku of the Day UK Table}). }
    \label{Nishio Human Cycles Classification}
    \rule[1ex]{\textwidth}{0.1pt}
\end{figure}

Figure \ref{Nishio Human Cycles Classification} shows the result: each of the 22 difficulty levels (across the 5 websites used in the Sudoku puzzle dataset) is mapped onto one of three universal categories using both its mean and median Nishio Human Cycles using 4 Human Strategies.

\paragraph{Mapping Difficulty Levels via Clause Length Distribution}
The same binning procedure is applied to the Clause Length Distribution, yielding three ranges:
\begin{itemize}
    \item Universal Easy: [100\%, 22.6 \%)
    \item Universal Medium: [22.6 \%, 17.6 \%)
    \item Universal Hard: [17.6 \%, 0 \%]
\end{itemize}

\begin{figure}[H]
    \centering
    \rule[1ex]{\textwidth}{0.1pt}
    \makebox[\textwidth][c]{\includegraphics[width=1.22\textwidth]{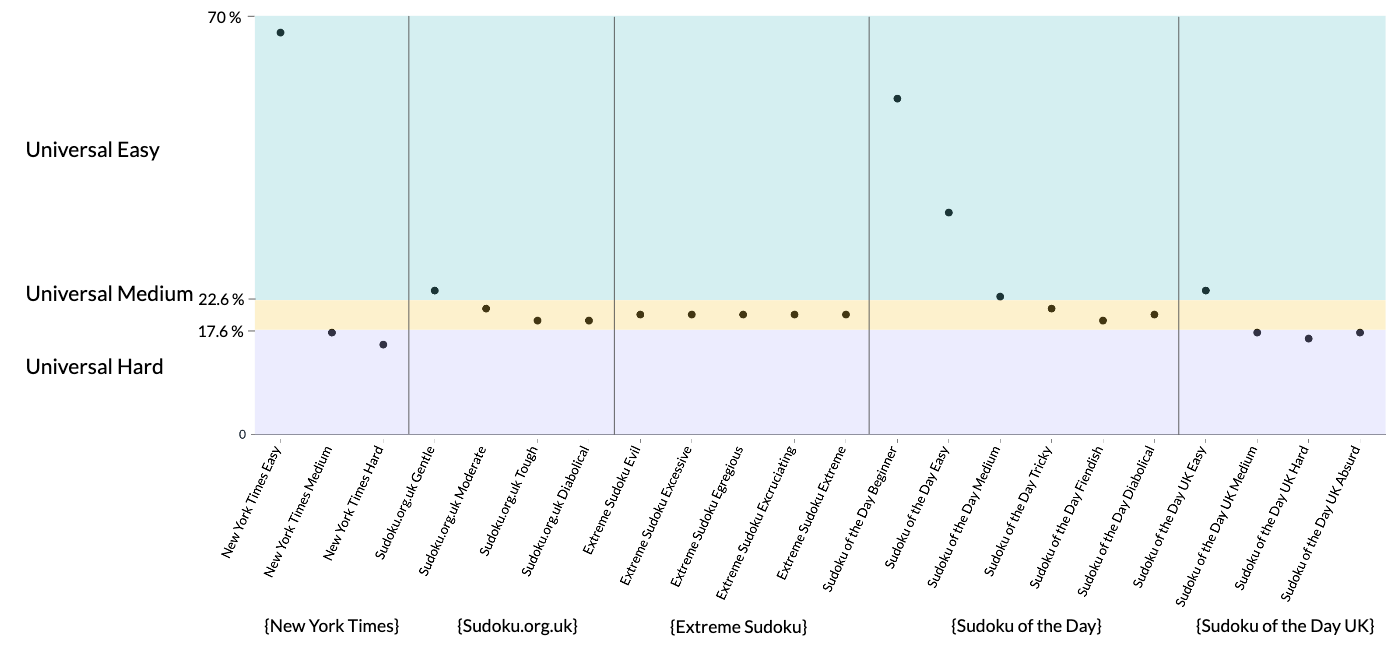}}%
    \caption{Difficulty levels from the 5 websites classified into one of three universal categories based on Mean percentage of short clauses (from Table \ref{SAT Table}). }
    \label{Clause Length Classification}
    \rule[1ex]{\textwidth}{0.1pt}
\end{figure}

Again, the universal classification maps 22 difficulty levels onto the three universal categories using the mean short-clause percentage values from Table \ref{SAT Table}. The resulting classification is shown in Figure \ref{Clause Length Classification}.

\subsection{Discussion and Further Analysis}
\subsubsection{Key Observations}
\label{Observations}
\begin{enumerate}
    \item Extreme Sudoku is anomalous, with all difficulty levels classifying as Universal Hard with mean Nishio Human Cycles and all classifying as Universal Medium with Clause Length Distribution.  Moreover, neither of the proposed metrics correlate with the website's assigned difficulty ordering (see Table \ref{Spearman's}). This anomalous behavior is further discussed in Section~\ref{Extreme Sudoku Anomaly}.
    \item In every difficulty level across all 5 websites, the mean Nishio Human Cycles is greater than the median Nishio Human Cycles, indicating a skewed right distribution. There are two possible reasons for this skewed right distribution \cite{skewness}: 
        \begin{itemize}
            \item There are outlier puzzlers that drag the data skewed right, i.e., the more difficult puzzles in each difficulty level are substantially more difficult than other puzzles. 
            \item  Individual Sudoku websites impose a lower bound on puzzle difficulty within each difficulty level,  intentionally generating puzzles such that they do not fall under that lower bound. Empirically, my data supports this conjecture since in all difficulty levels across all 5 websites, the lowest Nishio Human Cycles from a puzzle is lower than the lowest puzzle of all difficulty levels above it. 
        \end{itemize}
    \item 19 of the 22 difficulty levels fall into or near the "Universal Medium" difficulty level when classified by Clause Length Distribution. This trend indicates that Sudoku puzzles have little variation in clause length - specifically short clauses - regardless of their assigned difficulty. Therefore, a puzzle's Clause Length Distribution, capturing its structural complexity, including the number of givens, the values used, and the cells they are in, has minimal correlation with puzzle difficulty, as also shown in Table \ref{Spearman's}.
    
    \item 8 of the 22 difficulty levels are classified in a lower difficulty classification using Clause Length Distribution than classifying based on mean Nishio Human Cycles. Thus, we can surmise that the structure of a Sudoku puzzle demonstrates easier puzzle difficulty than the procedural difficulty in solving the Sudoku puzzle.  
    
\end{enumerate}

\subsubsection{Investigating the Anomalous Behavior in Extreme Sudoku}
\label{Extreme Sudoku Anomaly}
To better understand the anomalous behavior observed in the Extreme Sudoku dataset (see Section \ref{Extreme Sudoku} and Section \ref{Observations}), I focused my investigation on how Extreme Sudoku puzzles might be constructed. This investigation revealed that a large number of Extreme Sudoku puzzles, across different difficulty levels, are \emph{constructively identical}: a term I use to describe two or more puzzles that have the exact same given cells regardless of which  digits fill those cells. Figure \ref{fig:structural-anomaly} shows an example of three Extreme Sudoku puzzles that are constructively identical from three different difficulty levels. 

\newcommand{\givencells}{%
  1/3,1/4,1/9, 2/2,2/5,2/8, 3/1,3/6,3/7,
  4/1,4/6,4/7, 5/2,5/5,5/8, 6/3,6/4,6/9,
  7/3,7/4,7/9, 8/2,8/5,8/8, 9/1,9/6,9/7}

\newcommand{\maskgrid}[1]{%
  \edef\gclist{#1}%
  \begin{tikzpicture}[scale=0.38]
    \foreach \r/\c in \gclist {\fill[black!40] (\c-1,9-\r) rectangle ++(1,1);}
    \draw[step=1, black, thin]       (0,0) grid (9,9);
    \draw[step=3, black, very thick] (0,0) grid (9,9);
    \draw[black, very thick]         (0,0) rectangle (9,9);
  \end{tikzpicture}}

\begin{figure}[H]
  \centering
  \begin{subfigure}[t]{0.46\textwidth}
    \centering
    \maskgrid{\givencells}
    \caption{Egregious -- 01-04-25}
    \label{fig:mask-egregious}
    \vspace{0.3em}
    \footnotesize
    \begin{tabular}{@{}lr@{}}
      Nishio Human Cycles: & 5.02\\
      Short clauses:       & 21.30\,\%\\
      Medium clauses:      & 74.07\,\%\\
      Long clauses:        & 4.63\,\%\\
    \end{tabular}
  \end{subfigure}
  \hfill
  \begin{subfigure}[t]{0.46\textwidth}
    \centering
    \maskgrid{\givencells}
    \caption{Extreme -- 01-25-25}
    \label{fig:mask-extreme}
    \vspace{0.3em}
    \footnotesize
    \begin{tabular}{@{}lr@{}}
      Nishio Human Cycles: & 5.00\\
      Short clauses:       & 20.83\,\%\\
      Medium clauses:      & 74.07\,\%\\
      Long clauses:        & 5.09\,\%\\
    \end{tabular}
  \end{subfigure}

  \vspace{1.2em}

  \begin{subfigure}[t]{0.46\textwidth}
    \centering
    \maskgrid{\givencells}
    \caption{Excessive -- 01-25-25}
    \label{fig:mask-excessive}
    \vspace{0.3em}
    \footnotesize
    \begin{tabular}{@{}lr@{}}
      Nishio Human Cycles: & 4.36\\
      Short clauses:       & 20.83\,\%\\
      Medium clauses:      & 74.07\,\%\\
      Long clauses:        & 5.09\,\%\\
    \end{tabular}
  \end{subfigure}

  \caption{\textit{Three constructively identical Extreme Sudoku puzzles from three
    different difficulty levels: Egregious, Extreme, and Excessive. Given cells are
    shaded. Notice that the Nishio Human Cycles and Clause Length Distribution metrics
    are close across all three puzzles.}}
  \label{fig:structural-anomaly}
\end{figure}

Notice in Figure~\ref{fig:structural-anomaly} that a puzzle can be constructively identical to puzzles in more than one other difficulty level. In addition, a puzzle can be constructively identical to another puzzle in the same difficulty level. The exact pattern of given cells seen in Figure~\ref{fig:structural-anomaly} appears in 36 puzzles across all five difficulty levels from the 300-puzzle Extreme Sudoku dataset. 

Table \ref{tab:template-reuse} reports how often constructively identical puzzles occur within the Extreme Sudoku dataset. For instance, the first row in Table \ref{tab:template-reuse} shows that 44 out of 60 puzzles in the Evil difficulty level are constructively identical to at least one puzzle from at least one other different difficulty level. \emph{The total percentage of such constructively identical puzzles in the Extreme Sudoku dataset is 81\%. The same percentage  across the other 4 websites in this study is $\sim$0.6\%.} While this difference is dramatic, further investigation is needed to fully understand the extent to which constructively identical puzzles affect the difficulty labels assigned by Extreme Sudoku.

\begin{table}[H]
\centering
\captionsetup{width=.75\textwidth}
\caption{Constructively identical puzzle counts for each Extreme Sudoku difficulty level (out of 60).}
\label{tab:template-reuse}
\begin{tabular}{lc}
\toprule
Difficulty Level & Constructively Identical Puzzles  \\
\midrule
Evil          & 44/60 \\
Excessive     & 45/60 \\
Egregious     & 50/60 \\
Excruciating  & 54/60 \\
Extreme       & 49/60 \\
\bottomrule
\end{tabular}
\end{table}



\section{Heuristic-Based Nishio and Method for Early Sudoku Practitioners}
In Section \ref{simulating human solvers}, the solver in Figure \ref{randomized nishio} simulates human solvers, under assumptions of which strategies they know, using a randomization of Nishio. I now present a Nishio methodology that uses heuristics in place of randomization in step 2 in Figure \ref{randomized nishio}. While this method, shown in Figure \ref{deterministic nishio} is deterministic, it is not ideal for human use as it requires extensive bookkeeping to apply this method effectively. To address this, I simplify the method by adapting it for human Sudoku solvers, replacing the extensive bookkeeping with quick puzzle-scanning techniques.

\subsection{Heuristic-Based Nishio}
\begin{figure}[H]
    \centering
    \rule[1ex]{\textwidth}{0.1pt}
    \begin{enumerate}
    \item The candidates are computed for every cell as described in Section \ref{Running Example and Candidates}. 
    \item Unlike in step 2 of Figure \ref{randomized nishio} the solvers identifies a cell (called \textit{variable ordering}) and a candidate (called \textit{value ordering}) using the heuristics described below.
        \begin{itemize}
            \item \textbf{Variable Ordering} Two heuristics are used. \textit{Minimum remaining values}, selects the cell with the fewest number of candidates. If multiple cells have an equal number of candidates, the second heuristic, \textit{candidate counting} is applied. For each digit, 1-9, the number of occurrences as a candidate is counted. Then the sum of these occurrences for each candidate is computed for every cell.  For instance, in cell (1, 1) in Figure \ref{Easy Sudoku}, there are two candidates: 6 and 9. 6 appears 19 times and 9 appears as candidate 24 times, so the sum for cell (1, 1) is 43. Among cells with the least number of candidates, the cell with the highest sum is chosen first. Why? This heuristic prioritizes cells containing candidates that occur more frequently which results in more candidates being eliminated once a candidate is assumed, progressing the puzzle in fewer steps.
            \item \textbf{Value Ordering} Candidate counting is used for variable ordering. The candidate with the highest occurrence count is assumed first in the selected cell. For example, in cell (1, 1) in Figure \ref{Easy Sudoku}, the digit 9, which appears more often as a candidate than the digit 6, is assumed first in that cell. 
        \end{itemize}
    \item The four human strategies are applied in similar fashion to step 3 in Figure \ref{randomized nishio} until the puzzle cannot be progressed anymore.
    \item When puzzle progress halts, there are exactly three possible actions: 
        \begin{itemize}
            \item The puzzle is solved and no action is needed.
            \item There is at least one cell with no candidates left. In this case eliminate the candidate assumed in step 3 as a candidate and return to step 3.  
            \item The puzzle is not solved and there are not cells with no candidates, meaning another guess is needed.  In this case retain the current puzzle state and return to step 3.
        \end{itemize}
    \end{enumerate}
    \caption{Nishio using Heuristics}
    \label{deterministic nishio}
    \rule[1ex]{\textwidth}{0.1pt}
\end{figure}

The solver in Figure \ref{deterministic nishio} was run on the 1320-puzzle dataset, and the results for Nishio Human Cycles using both 2 and 4 strategies are summarized in Table \ref{deterministic}. Notably, the mean and median Human Cycles for every difficulty level across all 5 websites are lower than those produced using Randomized Nishio with Human Strategies (Figure \ref{randomized nishio}. This supports Nishio with heuristic-based variable and value ordering as a viable and more effective approach than using randomization , as implemented in Figure \ref{randomized nishio}.

\begin{table}[H]
\caption{Nishio Human Cycles (2 and 4 strategies) computed using the solver in Figure \ref{deterministic nishio} on the five-website dataset.} \label{deterministic}
\resizebox{\textwidth}{!}{
\begin{tabular}{lclccll|lllllcc}
\multicolumn{3}{l}{\multirow{2}{*}{}}                      & \multicolumn{2}{c}{Nishio Human Cycles}         &  &  &  &  &                   &                      & \multirow{2}{*}{} & \multicolumn{2}{c}{Nishio Human Cycles}                 \\ \cline{4-5} \cline{13-14} 
\multicolumn{3}{l}{}                                       & \multicolumn{1}{l}{}             &              &  &  &  &  &                   &                      &                   & \multicolumn{1}{l}{}             & \multicolumn{1}{l}{} \\
\multicolumn{3}{l}{Website}                                & \multicolumn{1}{l}{4 Strategies} & 2 Strategies &  &  &  &  & \multicolumn{3}{l}{Website}                                  & \multicolumn{1}{l}{4 Strategies} & 2 Strategies         \\ \hline
New York Times & \multicolumn{1}{l}{}             &        & \multicolumn{1}{l}{}             &              &  &  &  &  & Sudoku of the Day &                      &                   & \multicolumn{1}{l}{}             & \multicolumn{1}{l}{} \\
               & \multicolumn{1}{l}{Easy}         &        &                                  &              &  &  &  &  &                   & Beginner             &                   &                                  &                      \\
               &                                  & Mean   & 1                                & 1            &  &  &  &  &                   & \multicolumn{1}{c}{} & Mean              & 1                                & 1                    \\
               &                                  & Median & 1                                & 1            &  &  &  &  &                   & \multicolumn{1}{c}{} & Median            & 1                                & 1                    \\
               & \multicolumn{1}{l}{Medium}       &        &                                  &              &  &  &  &  &                   & Easy                 &                   &                                  &                      \\
               &                                  & Mean   & 4.1                              & 7.6          &  &  &  &  &                   & \multicolumn{1}{c}{} & Mean              & 1                                & 1                    \\
               &                                  & Median & 3                                & 6            &  &  &  &  &                   & \multicolumn{1}{c}{} & Median            & 1                                & 1                    \\
               & \multicolumn{1}{l}{Hard}         &        &                                  &              &  &  &  &  &                   & Medium               &                   &                                  &                      \\
               &                                  & Mean   & 8.05                             & 15.63        &  &  &  &  &                   & \multicolumn{1}{c}{} & Mean              & 2.42                             & 5.07                 \\
               &                                  & Median & 5.5                              & 12.5         &  &  &  &  &                   & \multicolumn{1}{c}{} & Median            & 2                                & 4                    \\
Sudoku.org.uk  & \multicolumn{1}{l}{}             &        & \multicolumn{1}{l}{}             &              &  &  &  &  &                   & Tricky               &                   &                                  &                      \\
               & \multicolumn{1}{l}{Gentle}       &        &                                  &              &  &  &  &  &                   & \multicolumn{1}{c}{} & Mean              & 4.82                             & 7.15                 \\
               &                                  & Mean   & 1.15                             & 1.15         &  &  &  &  &                   & \multicolumn{1}{c}{} & Median            & 4                                & 5.5                  \\
               &                                  & Median & 1                                & 1            &  &  &  &  &                   & Fiendish             &                   &                                  &                      \\
               & \multicolumn{1}{l}{Moderate}     &        &                                  &              &  &  &  &  &                   & \multicolumn{1}{c}{} & Mean              & 5.88                             & 8.25                 \\
               &                                  & Mean   & 1.8                              & 1.88         &  &  &  &  &                   & \multicolumn{1}{c}{} & Median            & 5                                & 7                    \\
               & \multicolumn{1}{l}{}             & Median & 1                                & 1            &  &  &  &  &                   & Diabolical           &                   &                                  &                      \\
               & \multicolumn{1}{l}{Tough}        &        &                                  &              &  &  &  &  &                   & \multicolumn{1}{c}{} & Mean              & 7.18                             & 7.63                 \\
               &                                  & Mean   & 4.67                             & 6.27         &  &  &  &  &                   & \multicolumn{1}{c}{} & Median            & 6                                & 6                    \\
               & \multicolumn{1}{l}{}             & Median & 4                                & 5            &  &  &  &  & Sudoku of the Day UK&                      &                   & \multicolumn{1}{l}{}             & \multicolumn{1}{l}{} \\
               & \multicolumn{1}{l}{Diabolical}   &        &                                  &              &  &  &  &  &                   & Easy                 &                   &                                  &                      \\
               &                                  & Mean   & 6.72                             & 6.7          &  &  &  &  &                   & \multicolumn{1}{c}{} & Mean              & 1.78                             & 2.43                 \\
               &                                  & Median & 5.5                              & 5.5          &  &  &  &  &                   & \multicolumn{1}{c}{} & Median            & 1                                & 1                    \\
Extreme Sudoku & \multicolumn{1}{l}{}             &        & \multicolumn{1}{l}{}             &              &  &  &  &  &                   & Medium               &                   &                                  &                      \\
               & \multicolumn{1}{l}{Evil}         &        &                                  &              &  &  &  &  &                   & \multicolumn{1}{c}{} & Mean              & 2.02                             & 2.53                 \\
               &                                  & Mean   & 6.17                             & 10.48        &  &  &  &  &                   & \multicolumn{1}{c}{} & Median            & 1.5                              & 1.5                  \\
               &                                  & Median & 5.5                              & 8            &  &  &  &  &                   & Hard                 &                   &                                  &                      \\
               & \multicolumn{1}{l}{Excessive}    &        &                                  &              &  &  &  &  &                   & \multicolumn{1}{c}{} & Mean              & 4.65                             & 7.63                 \\
               &                                  & Mean   & 6.7                              & 9.85         &  &  &  &  &                   & \multicolumn{1}{c}{} & Median            & 3                                & 4                    \\
               &                                  & Median & 6                                & 9            &  &  &  &  &                   & Absurd               &                   &                                  &                      \\
               & \multicolumn{1}{l}{Egregious}    &        &                                  &              &  &  &  &  &                   & \multicolumn{1}{c}{} & Mean              & 10.08                            & 11.1                 \\
               &                                  & Mean   & 5.9                              & 8.02         &  &  &  &  &                   & \multicolumn{1}{c}{} & Median            & 9.5                              & 10                   \\ \cline{8-14} 
               &                                  & Median & 5                                & 7            &  &  &  &  &                   &                      &                   & \multicolumn{1}{l}{}             & \multicolumn{1}{l}{} \\
               & \multicolumn{1}{l}{Excruciating} &        &                                  &              &  &  &  &  &                   &                      &                   & \multicolumn{1}{l}{}             & \multicolumn{1}{l}{} \\
               &                                  & Mean   & 6.68                             & 9.4          &  &  &  &  &                   &                      &                   & \multicolumn{1}{l}{}             & \multicolumn{1}{l}{} \\
               &                                  & Median & 6                                & 6.5          &  &  &  &  &                   &                      &                   & \multicolumn{1}{l}{}             & \multicolumn{1}{l}{} \\
               & \multicolumn{1}{l}{Extreme}      &        &                                  &              &  &  &  &  &                   &                      &                   & \multicolumn{1}{l}{}             & \multicolumn{1}{l}{} \\
               &                                  & Mean   & 6.35                             & 9.87         &  &  &  &  &                   &                      &                   & \multicolumn{1}{l}{}             & \multicolumn{1}{l}{} \\
               &                                  & Median & 5                                & 7            &  &  &  &  &                   &                      &                   & \multicolumn{1}{l}{}             & \multicolumn{1}{l}{} \\ \cline{1-7}
\end{tabular}
}
\end{table}

\subsection{A Method for Early Sudoku Practitioners to Solve Sudoku Puzzles} \label{human method}
The heuristic-based method described in Figure \ref{deterministic nishio}  requires extensive bookkeeping (steps 3 and 5). Specifically, there are two steps for which simple techniques are employed, including quick puzzle-scanning and judgment calls by the Hsolver. Additionally, the Hsolver should always apply the human strategies before using Nishio, as described in Step 1.5 below. 

\paragraph{Step 1.5: Application of Human Strategies Before Nishio} The Hsolver should always apply the 4 human strategies before trying Nishio. Nishio should only be used if the Hsolver is certain that none of the four strategies can advance the solution. A significant portion of Sudoku puzzles from the 1320-puzzle dataset can be solved using just these four human strategies, as seen in Table \ref{NyTimes Table} to Table \ref{Sudoku of the Day UK Table}.
\paragraph{Step 3: Candidate Counting} In step 2 of Figure \ref{deterministic nishio}, manually counting the occurrences of each digit (1-9) as a candidate in the puzzle is too impractical for an Hsolver. Instead, after applying minimum remaining values, the Hsolver should do a quick scan of the puzzle (2-3 seconds) to get a general sense of which digits appear most frequently as a candidate. Based on this, the Hsolver should pick a cell (from those containing the fewest candidates) that contains a candidate(s) that appears frequently. 

\paragraph{Step 4: Application of Human Strategies} After a candidate is assumed, the Hsolver applies the human strategies to make as much progress as possible (step 4 in Figure \ref{deterministic nishio}). However, instead of applying strategies until the puzzle is solved or there is a cell with no candidates, requiring tens of Nishio Human Cycles and multiple levels of assuming candidates (bullet 3 in step 4), the Hsolver should stop after 1 - 2 Nishio Human Cycles to avoid extensive bookkeeping.

At this point, if the puzzle is solved or there is a cell with no candidates, the Hsolver should continue with the next step in Figure \ref{deterministic nishio}. If neither happens, there are two options: 
\begin{itemize}
    \item Backtrack and assume a different candidate in the same cell. 
    \item Choose a new cell using the heuristics described in step 3 of Figure \ref{deterministic nishio} and assume another candidate, retaining the current state of the board. Keep in mind, this option should rarely be chosen, since it is too complex for the Hsolver to keep track of which values they have filled in for which level of backtracking in Nishio. 
\end{itemize}

 In testing on myself, using the method described in Section \ref{human method} led to noticeable improvements in my Sudoku solving times across Sudoku puzzles from all difficulty levels in each of the five websites. An area for future work is to conduct a larger-scale user study to evaluate the method’s effectiveness for early Sudoku practitioners. 

\section{Related Work} \label{related work}

\paragraph{Sudoku as a Satisfiability (SAT) problem}
There have been multiple efforts studying Sudoku as a Satisfiability problem \cite{chaos} \cite{kwon2006optimized} \cite{Sudoku-as-SAT} \cite{pfeiffer2010sudoku}. Closely related work is that of Ercsey-Ravasz and Toroczkai \cite{chaos}, who propose a scalar measure of Sudoku hardness: the \textit{escape rate} of a deterministic, continuous time dynamical system solver when solving a Sudoku puzzle's SAT instance. They compute the escape rate for individual Sudoku puzzles from 2 of the 5 websites used in this study, Extreme Sudoku and Sudoku of the Day UK. Specifically, Figure 4 of their paper presents 5 puzzles from Extreme Sudoku and 25 puzzles from Sudoku of the Day UK classified into three different categories based on escape rate: Easy, Medium, and Hard. For better readability, these three categories will be called Escape Rate Easy, Escape Rate Medium, and Escape Rate Hard in this subsection (with apologies to Ercsey-Ravasz and Toroczkai).  The use of three categories by Ercsey-Ravasz and Toroczkai \cite{chaos} also motivates the three-bin universal classification discussed in Section \ref{universal classification}. 

The universal classification using Clause Length Distribution is almost identical to the escape rate classification \cite{chaos} of Sudoku puzzles. This is expected as both metrics are derived from SAT encodings of Sudoku puzzles. Nishio Human Cycles classification, however, has some notable differences. Specifically, escape rate classifies all five Extreme Sudoku puzzles \footnote[3]{Ercsey-Ravasz and Toroczkai \cite{chaos} do not disclose from which difficulty level these puzzles are from} as Escape Rate Medium, while Nishio Human Cycles classification categorizes all five Extreme Sudoku difficulty levels as Universal Hard. For Sudoku of the Day UK, both methods agree on the Easy and Medium difficulty level classifications, but differ for the Hard and Absurd difficulty levels. Nishio Human Cycles classifies Hard and Absurd as Universal Hard, whereas the escape rate classification \cite{chaos} classifies all Hard and Absurd puzzles as Escape Rate Medium - except for one single puzzle, which is classified as Escape Rate Hard.

As discussed in Section \ref{spearman's section}, Clause Length Distribution and Nishio Human Cycles capture different aspects of Sudoku difficulty. Since both Clause Length Distribution and the escape rate \cite{chaos} are derived from Sudoku SAT encodings, both reflect the structural complexity of a Sudoku puzzle. Nishio Human Cycles produces consistently higher difficulty classifications, indicating that the structure of a Sudoku puzzle (including the number of givens, the values used, and the cells they are in) reflects easier puzzle difficulty than the procedural difficulty in solving the Sudoku puzzle. 

Ercsey-Ravasz and Toroczkai \cite{chaos} also have a fourth category in their classification, Ultra-Hard, reserved for a few challenging puzzles, described as the hardest puzzles in their analysis. The Nishio Human Cycles values for four of these puzzles are summarized in Table \ref{super hard}, all of which greatly exceed the 98.14 upper limit for the Universal Hard category range in Section \ref{universal classification}. This supports their description of these puzzles as the hardest Sudoku puzzles, classifying them into the Ultra-Hard category \cite{chaos}.

\begin{table}[H]
\caption{Nishio Human Cycles for 4 Ultra-Hard Sudoku puzzles}
\centering
\begin{tabular}{cl|lc}
Sudoku          &  &  & Nishio Human Cycles \\ \hline
Platinum Blonde &  &  & 1074.92      \\
Golden Nugget   &  &  & 750.22       \\
tarx0134        &  &  & 840.02       \\
Red Dwarf       &  &  & 508.88      
\end{tabular}

\label{super hard}
\end{table}

\paragraph{Constraint Propagation}
Constraint Programming (CP) is a foundational strategy in solving Constraint Satisfaction Problems (CSPs), and its application to Sudoku has been explored extensively \cite{constraint_programming1} \cite{constraint_programming2} \cite{constraint_prop} \cite{norvig}. One prominent example is Norvig \cite{norvig}, who combines constraint propagation with depth-first search to solve Sudoku puzzles. Specifically, Naked Singles and Hidden Singles are used in the algorithm written by Norvig \cite{norvig}, henceforth called Norvig's algorithm. I incorporate two additional, more sophisticated strategies, X-wing and Naked Twins, as described in Section \ref{strategies}. In Norvig's algorithm, Naked Singles and Hidden Singles, the constraint propagation techniques, are applied until the puzzle cannot be progressed anymore. Step 3 in Figure \ref{deterministic nishio}, is different from Norvig's algorithm. Variable ordering is done by using \textit{minimum remaining values}, with ties broken randomly. For value ordering, Norvig \cite{norvig} chooses the candidates in numeric order. In Norvig's algorithm, the use depth-first search is identical to steps 4 and 5 in Figure \ref{deterministic nishio}.

Crawford et al. \cite{constraint_programming2} evaluate multiple variable and value ordering heuristics in solving Sudoku using constraint programming. They find that \textit{minimum remaining values} outperforms \textit{maximum remaining values}, supporting the choice for minimum remaining values in Figure \ref{deterministic nishio}. While they test value ordering heuristics like \textit{smaller value of the domain}, Figure \ref{deterministic nishio} employs \textit{candidate counting} as a heuristic.

Simonis \cite{constraint_prop} involves evaluates the effectiveness of  combining multiple combinations of constraint programming techniques for Sudoku solving and rating Sudoku difficulty. Specifically, he reports the percentage of puzzles solved in each difficulty level from multiple websites. This metric is similar to the percentage of puzzles solved by only human strategies used in the within site analysis in Section \ref{within site analysis}. Unfortunately, none of the websites used by Simonis \cite{constraint_prop} overlap with the 5 websites used in the 1320-puzzle dataset.

\paragraph{Using Human Clock Time Solving to Determine Sudoku Difficulty}
Pelánek \cite{difficulty_rating} proposes various difficulty metrics derived from four distinct approaches: algorithm-based searches (backtracking, simulated annealing, and harmony search), puzzle given digits, constraint relaxation \cite{constraint_relaxation}, and a human-solving model that, at each step, applies the simplest strategy capable of making progress on the current puzzle state. The viability of these metrics are evaluated against mean human solve clock times, as recorded on Sudoku websites, including Sudoku.org.uk. However, as discussed in Pelánek’s  study \cite{difficulty_rating}, human solve time recorded by websites has many possible distortions. Since this data is not collected in a controlled experiment, human solvers may pause, take breaks, and return to solving later, leading to inflated solve times. Additionally, human solvers improve over time, and Pelánek \cite{difficulty_rating} shows a positive correlation between the number of days since a human solver's first puzzle and their current solve time. In contrast, I use difficulty levels provided by websites as a benchmark as it serves as a consistent measure of difficulty rating.

\paragraph{Sudoku as a Benchmark for Large Language Models}
An emerging line of research explores the use of Sudoku as a benchmark for evaluating and training logical reasoning in Large Language Models (LLMs) \cite{LLMs3} \cite{LLMs2} \cite{sudokubench} \cite{LLMs}. These efforts typically vary problem difficulty by altering puzzle sizes - such as using smaller grids like 4×4 or 6×6 and even using Sudoku variants. In contrast, my work focuses on the standard 9×9 Sudoku, which itself exhibits substantial variability in difficulty. Of particular relevance to this work is SudokuBench \cite{sudokubench}, which uses an unrated benchmark dataset, nikoli\textunderscore100, specifically curated to evaluate multi-step logical reasoning in LLMs. I applied the universal classification built with Nishio Human Cycles to this dataset. The resulting distribution, shown in Table \ref{nikoli_100}, highlights that nikoli\textunderscore100 contains notably simpler puzzles on average than those from the 5 Sudoku websites used in this study, with more than 50 \% of puzzles classified as Universal Easy. Such rated Sudoku datasets, which have a defined variability in difficulty, can be leveraged to train and evaluate reasoning in LLMs on top of varying the Sudoku size and type. A future area of research is investigating the use of the Randomized Nishio with Human Strategies solver described in Figure \ref{randomized nishio} to generate reasoning traces to improve the logical reasoning capabilities of LLMs. 
\begin{table}[H]
\centering
\caption{Universal Classification via Nishio Human Cycles for the nikoli_100 dataset} \label{nikoli_100}
\begin{tabular}{ll|lc}
Universal Category &  &  & \begin{tabular}[c]{@{}c@{}}Number of Puzzles \\ from nikoli\_100 dataset\end{tabular} \\ \hline
Universal Easy     &  &  & 53                                                                                    \\
Universal Medium   &  &  & 24                                                                                    \\
Universal Hard     &  &  & 23                                                                                   
\end{tabular}
\end{table}

\section{Conclusion}
In this study, we addressed the question of what constitutes Sudoku difficulty across 5 Sudoku websites. I present and evaluate two different methods to solve Sudoku puzzles: one that encodes Sudoku puzzles into Satisfiability (SAT) problems and the second that simulates a human solver by interleaving four popular human strategies within a backtracking method called Nishio. From these two methods, I derive two metrics for measuring puzzle difficulty. The first metric, Clause Length Distribution, is based on SAT encodings and captures the structural complexity of a Sudoku puzzle. The second metric, Nishio Human Cycles, captures the procedural difficulty in solving a Sudoku puzzle by counting how many human strategy cycles in the backtracking process are needed to solve the Sudoku puzzle. Using these metrics, I analyze and compare difficulty levels across 1320 puzzles from 5 popular Sudoku websites.

I analyze each website using the two proposed metrics: Clause Length Distribution and Nishio Human Cycles. The experimental results show that in four out of 5 websites, the labeled difficulty levels align with the two difficulty metrics. The exception is Extreme Sudoku, which exhibits anomalous patterns and does not correlate with any of the proposed difficulty metrics. To provide a unified view of difficulty ratings across websites, I introduce a simple, unsupervised classification, applied separately to each proposed metric. This method partitions puzzles into three difficulty categories: Universal Easy, Universal Medium, and Universal Hard. This universal classification enables consistent comparison across websites and supports classification of previously unlabeled datasets, such as nikoli\textunderscore100 from SudokuBench \cite{sudokubench}.

In the future I hope to continue to extend this work by using the methods described in Figure \ref{randomized nishio} and \ref{deterministic nishio} to generate reasoning traces to train and evaluate logical reasoning in large language models (LLMs) and introduce puzzle difficulty variations in datasets like nikoli\textunderscore100.

\section{Acknowledgments}
I would like to thank my grandmother, Shymala Vaidyanathan, for inspiring me to write this paper by asking the question "Why can I solve Diabolical puzzles on one Sudoku website but not Easy puzzles on another Sudoku website?"

Words cannot express my gratitude towards my Dad, who sat through tough technical discussions with me, offered feedback and guidance, and supported me through every step of the way. 

Special thanks to Rajasekar Krishnamurthy, a mentor who offered ideas during the early stages of the research, read an unfinished early draft, and provided helpful feedback throughout this journey.

I would like to thank the anonymous reviewer for encouraging the analysis presented in Section \ref{Extreme Sudoku Anomaly}.

Finally, I would like to thank the puzzle creators from Sudoku.org.uk, Extreme Sudoku, and Sudoku of the Day for granting me permission to publicly release their datasets for academic purposes. 

\nolinenumbers

\bibliographystyle{plain} 
\bibliography{references} 

\end{document}